\documentclass[11pt]{article}
\usepackage{amsmath,amsfonts}
\usepackage{algorithm}
\usepackage{algpseudocode}
\usepackage{times}
\usepackage{graphicx}
\usepackage{multirow}

\usepackage{bm}
\usepackage{amssymb}

\usepackage{rotating}
\usepackage{cite}
\usepackage{bbm}
\usepackage{latexsym}
\usepackage{fancyhdr}

\newcommand{\argmin}{\mathrm{ argmin}}

\begin{document}
\hspace{13.9cm}1

\ \vspace{20mm}\\

{\begin{center}
{\LARGE Double Sparse Multi-Frame Image Super Resolution}

\ \\
{\bf \large Toshiyuki Kato$^{1}$, Hideitsu Hino$^{2}$, and Noboru
  Murata$^{1}$} \\
$^{1}$Waseda University, 3-4-1 Ohkubo, Shinjuku,
Tokyo, Japan,\\
$^{2}$University of Tsukuba.\\
1-1-1 Tennodai, Tsukuba, Ibaraki, 305--8573, Japan
\end{center}}

\thispagestyle{fancy}
\rhead{}
\lhead{}

%Abstract
\begin{center} {\bf Abstract} \end{center}
A large number of image super resolution algorithms based on the sparse
coding are proposed, and some algorithms realize the multi-frame super
resolution. In multi-frame super resolution based on the sparse coding,
both accurate image registration and sparse coding are
required. Previous study on multi-frame super resolution based on sparse
coding firstly apply block matching for image registration, followed by
sparse coding to enhance the image resolution. In this paper, these two
problems are solved by optimizing a single objective function. The
results of numerical experiments support the effectiveness of the
proposed approch.
%%%%%%%%%%%

\section{Introduction}
\label{sec1}
Image super resolution (SR) is a problem of enhancing
the resolution of observed low resolution (LR) images. 
The importance of super resolution is increasing because of the growing
needs for remastering old films or investigating low resolution
surveillance videos, for example. A large number of methods are proposed for SR, and
a group of actively studied methods are based on sparse
coding~\cite{OLS96,OLS05,Nat1995,elad2010sparse}, which is the focus of
this paper. 
Sparse coding is a methodology in signal processing, where an observed
signal is approximated by a linear combination of simple components
called {\it{atoms}}. 
A distinctive feature of sparse coding is that the number of atoms
prepared for the signal reconstruction is large, while the number of
atoms actually used for representing a signal is small. The nature of
sparse coding enables us to compactly represent signals and effectively
remove noises.

Sparse coding is used for super resolution from single LR
image~\cite{yang1,Yang:2010:ISV:1892456.1892463,elad1} and recently
from multiple LR images~\cite{Kato2015}. 
 Multi-frame image SR is
expected to offer clearer high resolution (HR) image than single-frame
SR, if the relative position of observed LR images are accurately
estimated. Indeed, estimation of relative displacement or shift of
multiple images, which is referred to as {\it{image
registration}}~\cite{Zitov2003977}, plays a critical role in SR as well
as sparse coding. 

In our previous work for image SR~\cite{Kato2015}, we treated problems
of image registration and sparse coding separately. That is,
firstly we estimate relative displacement of observed LR images,
then we applied the sparse coding algorithm to aligned LR
images. Since the objective of registration of LR images is in realizing
 high-quality HR image restoration by sparse representation, it is natural
to perform image registration so that the error in sparse image
representation is minimized. The contribution of this paper is treating
the sub-pixel level image registration and sparse coding problems in a
unified framework. More concretely, we simultaneously estimate both
displacements of LR images and coefficients of SC with a single objective.
 Theoretically, we cast the multi-frame SR problem into a particular
 framework called {\it{double sparsity}}, which is an interesting
 approach for sparse modeling~\cite{rubinstein,DBLP:conf/icip/ZhanZYHH13}.

The rest of this paper is organized as follows. Section~2 describes the
problem setting and the underlying model of multi-frame super
resolution. The sparse coding approach for super
resolution is also shown in this section. 
Section~III briefly explains how fine relative displacements (shifts) between
observations are expressed by combinations of pixel-level displacements. 
Section~IV describes our
proposed approach for estimating displacements and sparse
coding coefficients in a unified framework. Section~V shows the experimental
results, and the last section is devoted to concluding remarks.

%%%%%%%%%
\section{Notation and Formulation}
We first explain the notion of single-frame super resolution by
sparse coding, then extend it to multi-frame super resolution.

\subsection{Single Frame Super Resolution}
\label{sec:SFSR}
Let $\mathbf{Y} \in \mathbb{R}^{Q}$ be the observed LR image. The
aim of super solution is constructing an HR image $\mathbf{X} \in
\mathbb{R}^{P}$ from $\mathbf{Y}$.
To reduce computational costs, image super
resolution is often performed for small image regions called {\it{patches}}, then
they are combined to construct a whole image. Following this way, we 
consider reconstructing an HR image patch represented by a vector
${\bf{x}} \in \mathbb{R}^{p}$ by using a single LR image patch ${\bf{y}} \in \mathbb{R}^{q}$.
 After obtaining all the HR patches, certain post-processing
for constructing the full-size HR image is performed, which is explained
in sections 5.2 and 5.3 of our previous paper in detail~\cite{Kato2015}. 

For each patch, we assume the following degradation process:
each patch pair
$(\mathbf{x}, \mathbf{y})$ is connected by the observation model
\begin{align}
\label{eq:Smodel_patch}
\mathbf{y} = G\mathbf{x} + \bm{\varepsilon},
\end{align}
where $G$ is a degradation operator composed of blur and down-sampling
operations, and $\bm{\varepsilon} \in \mathbb{R}^{q}$ is the additive
observation noise. 

Sparse coding~\cite{OLS96,elad2010sparse} is a methodology to represent observed signals with combinations
of only a small number of basis vectors chosen from a large number of candidates.
These basis vectors will be called {\it atoms} henceforth. 

Let $\mathbf{D} = [\mathbf{d}_1,\mathbf{d}_2,\dots,\mathbf{d}_K] \in
\mathbb{R}^{q \times K}$ be a {\it{dictionary}} which consists of $K$
atoms, and let ${\boldsymbol \alpha \in \mathbb{R}^{K}}$ be the coefficient
vector for sparse representation of the patch $\mathbf{y} \in
\mathbb{R}^{q}$. Typically, $K > q$. The problem of sparse coding is
formulated as follows:
\begin{align}
\underset{{{\boldsymbol{\alpha}}}}{\text{minimize}}
  \| {\mathbf{y}}- \mathbf{D}{\boldsymbol
 \alpha}\|_2^2 + \eta
\|{\boldsymbol \alpha}\|_1, \quad \eta >0,
\label{eq:bp}
\end{align}
where $\|{\boldsymbol \alpha}\|_{1} = \sum_{j=1}^{K} |\alpha_{j}|$ is
the $\ell^{1}$-norm of a vector. 
This problem~\eqref{eq:bp} adopts the $\ell^1$-norm of coefficients as a
measure of sparsity, and is referred to as the $\ell^1$-norm sparse
coding. By minimizing both the approximation error and the $\ell^{1}$-norm of the coefficient of the
atoms for patch representation, the resultant coefficient $\boldsymbol
\alpha$ has only a few nonzero elements, and the observed patch
$\mathbf{y}$ is well approximated by using a small number of atoms.
 More specifically, we call the problem of obtaining the coefficients
${\boldsymbol \alpha}$ with a fixed dictionary $\mathbf{D}$ {\it{sparse
coding}}. 

On the other hand, the problem of optimizing the dictionary
${\mathbf{D}}$ with a set of observations
$\{{\mathbf{y}}_{i}\}_{i=1}^{n}$ is called {\it{dictionary
learning}}:
\begin{align}
\underset{{{\mathbf{D}}} }{\text{minimize}}
\sum_{i=1}^{n}
 \| {\mathbf{y}_{i}}- \mathbf{D}{\boldsymbol
 \alpha}_{i}\|_2^2,
\end{align}
where ${\boldsymbol{\alpha}}_{i}$ is given by solving the
problem~\eqref{eq:bp} for each ${\mathbf{y}}_{i}$. 
 There are a number of methods for dictionary
 learning~\cite{Engan:1999:MOD:1257298.1257971,ksvd} and
 sparse
 coding~\cite{Rezaiifar93orthogonalmatching,Tibshirani94regressionshrinkage}. In
 this work, we use algorithms for learning dictionary and coefficients
 proposed in~\cite{DBLP:conf/nips/LeeBRN06} because of
 their computational efficiency. 

\begin{figure}[!t]
 \centering
 \includegraphics[scale=.37]{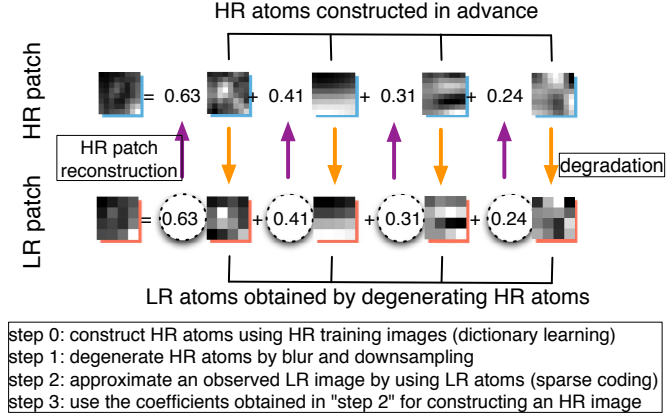}
 \caption{An illustrative diagram of super resolution by sparse coding. 
\label{fig:SCSR}}
 \end{figure}

We assume the HR patch $\bf{x}$ is represented by a sparse combination
of HR atoms as ${\bf{x}} = {\mathbf{D}}^{h} \boldsymbol \alpha$. Because
of relation \eqref{eq:Smodel_patch}, the LR patch ${\bf{y}}$ is
connected with the HR patch ${\bf{x}}$ as
\begin{align}
\mathbf{y} \simeq G {\mathbf{x}} = G \mathbf{D}^{h} {\boldsymbol \alpha} =
 {\mathbf{D}^{l}} {\boldsymbol \alpha},
\label{eq_scsr}
\end{align}
where $\mathbf{D}^{l}$ is the LR dictionary, atoms in which are
generated from $\mathbf{D}^{h}$ by the above explained image degeneration
process and have one-to-one correspondence to the HR atoms. 
The above correspondence \eqref{eq_scsr} naturally leads us to the
following two step procedure for single-frame super resolution: firstly representing the LR patch by the
sparse combination of LR atoms, and secondly reconstructing the
corresponding HR patch by using the combination coefficients for
the LR atoms for combining the HR atoms as shown in
Fig.~\ref{fig:SCSR}. 
Before performing this single-frame SR, an HR dictionary has to be
prepared by an appropriate dictionary learning algorithm with HR
training images.

\subsection{Multi-frame Super Resolution by Sparse Coding}
\label{sec:MFSR}
Suppose $N$ low resolution images $\mathbf{Y}_{i} \in \mathbb{R}^{Q},
i=1,\dots, N$ are observed, all of which are differently degenerated
from a high resolution image $\mathbf{X} \in \mathbb{R}^{P}$. 
 Without loss of generality, we assume that the LR image
 $\mathbf{Y}_{1}$ is the {\it{target}} for super resolution, and other
 $N-1$ LR images are called {\it{auxiliary}} images. 
 
We assume that each patch is exposed by the following
image degradation process: for the target patch ${\mathbf{y}}_{1}$, the
image degradation is modeled by Eq.~\eqref{eq:Smodel_patch}, and for
auxiliary patch pairs $(\mathbf{x}, \mathbf{y}_{i}), i=2,\dots, N$,
observation model 
\begin{align}
 \label{eq:Mmodel_patch}
 \mathbf{y}_{i} =  G W_{i}\mathbf{x} +
 \bm{\varepsilon}
\end{align}
is assumed, where $W_{i}$ is the parallel shift and clipping operator
corresponds to the $i$-th LR patch, which is explained later.
We note that $G$ can be not identical for different images
${\mathbf{Y}}_{i}$, but they are assumed to be identical for
the sake of simplicity, and we concentrate on estimating $W_{i}, i=2,\dots,N$ for $N-1$ different observations.

In our multi-frame SR, firstly, the target patch ${\bf{y}}_{1}$ is extracted from the target
image $\mathbf{Y}_{1}$. 
Then we consider estimating the shift of the
$i$-th image $\mathbf{Y}_{i}$. 
We can roughly estimate the displacement of the target patch
${\bf{y}}_{1}$ in the auxiliary image $\mathbf{Y}_{i}$
by sub-pixel-level accuracy matching in the LR space. 
 To avoid the negative effect at the non-overlapping
region and to use the informative area only, the pixels completely
included in the placed target patch is extracted as the $i$-th LR patch
${\bf{y}}_{i} \in \mathbb{R}^{(5-1)\times (5-1)}$. 
 Since the target patch ${\mathbf{y}}_{1}$ is represented by
a vector of length $q = \sqrt{q} \times \sqrt{q}$, the size of patch in
$2$-dimensional expression is $\sqrt{q} \times \sqrt{q}$. On the other
hand, the boundary-clipped patch is of size $q^{\prime} = (\sqrt{q}-1)
\times (\sqrt{q}-1)$. 

Once $W_{i},i=2,\dots,N$ are estimated, the image SR based on sparse
coding is straight-forward. 
With the shift and clipping $\{W_{2},\dots,W_{N}\}$ of
auxiliary patches ${\bf{y}}_{2},\dots,{\bf{y}}_{N} \in
\mathbb{R}^{q^{\prime}}$, the LR dictionaries correspond to the
observations $\{{\bf{y}}_{i}\}_{i=1}^{N}$
are stacked to construct a stacked LR dictionary:
\begin{align}
\label{eq:joint}
 \tilde{\mathbf{D}}^l &= \begin{bmatrix}
                          \mathbf{D}^l \\
                          \mathbf{D}^l_2 \\[3pt]
                          \mathbf{D}^l_3 \\
                          \vdots \\
                          \mathbf{D}^l_N
                         \end{bmatrix}
 =
                        \begin{bmatrix}
                         G\mathbf{D}^h \\
						 G W_2\mathbf{D}^h \\
                         G W_3\mathbf{D}^h \\
                         \vdots \\
						 G W_{N}\mathbf{D}^h
                        \end{bmatrix}.
\end{align}
This dictionary is used to approximate the stacked LR patch
\begin{equation}
\label{eq:jointy}
\tilde{\bf{y}} = 
\begin{bmatrix}
{\bf{y}}_{1}\\
{\bf{y}}_{2}\\
\vdots\\
{\bf{y}}_{N}
\end{bmatrix}
\end{equation}
by sparse coding, namely, the HR estimate $\hat{\bf{x}}$  is given by
\begin{align}
 \hat{\boldsymbol \alpha} &= \underset{{\boldsymbol \alpha}}{\argmin}
\|\hat{\bf{y}} - \tilde{\mathbf{D}}^{l} {\boldsymbol \alpha}\|_{2}^{2} +
 \eta \|{\boldsymbol \alpha}\|_{1},\\
\hat{\bf{x}} &=
{\bf{D}}^{h} \hat{\boldsymbol \alpha}.
\end{align}
Each atom ${\boldsymbol d}^{h}$ in the HR dictionary $\mathbf{D}^{h}$ is
shifted and clipped to an atom of size $q^{\prime}$ by the action of
$W_{i}, i=2,\dots,N$, then blurred,
down-sampled by the action of $G$ to form corresponding LR atoms
$G W_{i} {\boldsymbol d}^{h},i=2,\dots,N$. 
 Each block of the stacked
dictionary in Eq.~\eqref{eq:joint} is composed of LR atoms obtained in
this manner.

\section{Approximation of Displacements by Pixel-Level Shifts}
As discussed in the previous section, the main problem of multi-frame SR
is reduced to the problem of estimating shift and clipping operators
$W_{i}, i=2,\dots,N$. 
In the following, we consider enhancing the resolution of the
magnification factor $k$. Since the LR target
patch ${\mathbf{y}}_{1} \in \mathbb{R}^{q}$ is represented by a square
with $\sqrt{q}$ pixels on a side, the corresponding 
 HR patch is a square with
$\sqrt{p} =k \sqrt{q}$ pixels, namely, a vector of size $p=k^2 q$.

For estimating $W_{i}, i=2,\dots,N$,
we consider the upper left most point of the $i$-th auxiliary patch
${\bf{y}}_{i}$ as the origin $(0,0)$ of the shift, which is on the grid
of the $i$-th LR image $\mathbf{Y}_{i}$. 
We consider the displacement of the target LR patch with HR accuracy in
order to achieve satisfactory result, that means placing
the target LR patch on the LR image $\mathbf{Y}_{i}$ by using sub-pixel level matching.
Let the upper left nearest grid point from the origin in the $i$-th LR image
$\mathbf{Y}_{i}$ be $(k,k)$, and the upper left most
point of the placed target patch be $(a,b) \in [0,k]^{2}$.

There are two cases of displacement: pixel-level parallel shifts in the
HR image space, and others.
In the former case where $(a,b)$ are both integers, the LR dictionary
for the auxiliary patch can be simply constructed by clipping the
corresponding areas from the HR dictionary and degradation with $G$. 
Those LR dictionaries of pixel-level parallel shifts $a,b = 0,1,\dots,k$
are denoted by ${\mathbf{D}}^{l(a,b)}$, and referred to as the {\it{LR
base dictionaries}} henceforth. In the latter case, we assume that
relative displacement of the auxiliary patch is well-approximated by a linear combination of
pixel-level parallel translation in the HR space. Namely, estimating
shift and clipping operators is reduced to estimating the combination
coefficients for possible $(k+1)^{2}$ parallel translations. The $i$-th
block of the LR dictionary ${\mathbf{D}}^{l}_{i}$ in
Eq.~\eqref{eq:joint} is then given by 
\begin{align}
\notag
{\mathbf{D}}^{l}_{i}
&= 
G W_{i} {\mathbf{D}}^{h} \\ \notag
 \label{eq:Dlelem}
 & = 
\theta_{i,(0,0)}\mathbf{D}^{l(0,0)} +
 \theta_{i,(0,1)}\mathbf{D}^{l(0,1)} + \cdots +
 \theta_{i,(k,k)}\mathbf{D}^{l(k,k)}
\end{align}
where the LR base dictionaries are common to all of LR observations, and
different observation is represented by different coefficients $\theta$. 

In our
previous work~\cite{Kato2015}, we took a 2-step procedure for
multi-frame SR. That is, firstly we estimate the parameters
$\theta_{i,(a,b)}$ by using the 2D simultaneous block matching method
proposed in~\cite{2dest2} because of its
computational efficiency. Then, we construct the stacked LR dictionary in
Eq.~\eqref{eq:joint} and stacked LR patches in Eq.~\eqref{eq:jointy}, 
and the HR patch is obtained by sparse coding.
In this work, instead of the 2-step procedure, we propose a novel
approach for estimating the sub-pixel level accuracy shifts through a
common optimization objective to spares coding. 

\section{Double Sparsity for Image Super Resolution}
In this section, we formalize the proposed method for estimating both
displacements of LR images and coefficients for sparse coding.

\subsection{Problem Formulation}
We start with showing that the stacked LR patches $\tilde{\bf{y}}$ in
Eq.\eqref{eq:jointy} is approximated by a bi-linear form. This is done
by expressing the stacked dictionary in
Eq.~\eqref{eq:joint} as
\begin{equation}
\label{eq:DtildetI}
 \tilde{\mathbf{D}}^{l} 
 = {\mathbf{B}} \left( {\boldsymbol \theta} \otimes \mathbf{I}
					   \right),
\end{equation}
where $\otimes$ is the Kronecker product, $\mathbf{B}$ is defined by 
\begin{displaymath}
{\scriptscriptstyle
\begin{bmatrix}
 \mathbf{D}^l_{1}  & & & & && \\
  & \mathbf{D}^{l(0,0)} &  \cdots & \mathbf{D}^{l(k,k)} & & & & & \\
   & & & & \ddots & &  \\
   & & & & & \mathbf{D}^{l(0,0)} &  & \cdots & \mathbf{D}^{l(k,k)} \\
                                                 \end{bmatrix}
},
\end{displaymath}
$\boldsymbol \theta$ is defined by 
\begin{displaymath}
\begin{bmatrix} 1 , \theta_{2,(0,0)} , \theta_{2,(0,1)} , \hdots , \theta_{2,(k,k)} , \hdots, \theta_{N,(0,0)} , \hdots , \theta_{N,(k,k)}  \end{bmatrix}^{\top},
\end{displaymath}
and ${\mathbf{I}}$ is the unit matrix of an appropriate size.

Then, the approximation of an LR patch is denoted by
\begin{align}
 \tilde{\mathbf{y}} \simeq \tilde{\mathbf{D}}^{l} {\boldsymbol \alpha} 
&=  {\mathbf{B}} \left( {\boldsymbol \theta} \otimes \mathbf{I} \right)  {\boldsymbol \alpha} \\
&=   {\mathbf{B}} \left( {\mathbf{I}} \otimes {\boldsymbol \alpha}\right) {\boldsymbol \theta} \\
& = \mathbf{B}\; \mathrm{vec}\left( {\boldsymbol
								   \alpha}{\boldsymbol \theta}^\top
								  \right),
\end{align}
where $\mathrm{vec}$ is the column-span vectorization operator.
 In the above expression, ${\boldsymbol \alpha}$ is the coefficient
vector of sparse coding, and ${\boldsymbol \theta}$ is the shift
 vector of LR observations with sub-pixel level shifts in the HR space.
We will
estimate ${\boldsymbol \alpha}$ and ${\boldsymbol \theta}$ from
the observed LR patches. Then, the estimated coefficient ${\boldsymbol
\alpha}$ is used for reconstructing the HR image as $\mathbf{x} =
\mathbf{D}^h {\boldsymbol \alpha}$. 

The optimization problem to be solved is
\begin{align}
 \label{eq:objfunc}
\min_{\left\{ \boldsymbol \alpha , {\boldsymbol \theta} \right\}}\;
 \| \tilde{\mathbf{y}} - {\mathbf{B}} \; \mathrm{vec} \left( {\boldsymbol
 \alpha} {\boldsymbol \theta}^\top \right) \|_2^2 + \eta \|
 {\boldsymbol \alpha} \|_1 \\
 \text{subject to} \hspace{10pt} \mathbf{E}{\boldsymbol \theta} \leq \mathbf{1}, \hspace{10pt} {\boldsymbol \theta} \geq \mathbf{0},
\end{align}
where $\mathbf{1}$ denote the vector of all ones. 
The regularization term $\eta \| {\boldsymbol \alpha}\|_1$ imposes
sparsity for representing observed signals by linear combinations of atoms.
The inequality $\mathbf{E}{\boldsymbol \theta} \leq \mathbf{1}$ encodes
a constraint that the sum of coefficients for interpolation is less than
or equal to one for each image. Here the matrix 
$\mathbf{E} \in \mathbb{R}^{N \times \{1+(k+1)^{2}\times (N-1)\}}$ is defined by
\begin{align}
 \mathbf{E} = \begin{bmatrix}
 1 & & & & & & \multicolumn{3}{c}{\multirow{3}{*}{{\Huge 0}}} & \\
 & 1 & 1 & \cdots & 1 & & & & & \\
 & \multicolumn{3}{c}{\multirow{2}{*}{{\Huge 0}}} & & \ddots & & & & \\
 & & & & & & 1 & 1 & \cdots & 1 \\
 \end{bmatrix}.
\end{align}
Together with the non-negativity constraint ${\boldsymbol \theta} \geq
\mathbf{0}$, the constraints $\mathbf{E}{\boldsymbol \theta} \leq
\mathbf{1}$ and ${\boldsymbol \theta} \geq \mathbf{0}$ constitute the
$\ell^{1}$-norm like constraints with non-negativity, which also produces
sparse solutions not only for ${\boldsymbol \alpha}$ but also for ${\boldsymbol \theta}$.

\subsection{Optimization method}
\label{subsec:optim}
We solve the optimization problem~\eqref{eq:objfunc}. Since it is intractable to find a closed-form solution
for the problem~\eqref{eq:objfunc} on both ${\boldsymbol \alpha}$ and
${\boldsymbol \theta}$, we alternatingly solve the problem with respect to
${\boldsymbol \alpha}$ with a fixed ${\boldsymbol \theta}$ and with
respect to ${\boldsymbol \theta}$ with a fixed ${\boldsymbol \alpha}$. 

First of all, using only the target LR patch $\mathbf{y}_1$, we
initialize the coefficient ${\boldsymbol \alpha}^{(1)}$ by solving the
following optimization problem,
\begin{align}
 \label{eq:a0est}
 {\boldsymbol \alpha}^{(1)} &= \arg \min_{\boldsymbol \alpha} \| \mathbf{y}_1 - \mathbf{D}^{l} {\boldsymbol \alpha}\|_2^2 + \eta \| {\boldsymbol \alpha }\|_1,
\end{align}
which is efficiently solved by using sparse coding algorithms.

By fixing the coefficient ${\boldsymbol \alpha}^{(t)}$, we obtain the
combination coefficient for shift operators by 
\begin{equation}
\begin{split}
\label{eq:QP}
 {\boldsymbol \theta}^{(t)} & = \arg \min_{\boldsymbol \theta} \|
 \tilde{\mathbf{y}} - {\mathbf{B}} (\mathbf{I} \otimes
 {\boldsymbol \alpha}^{(t)}) {\boldsymbol \theta} \|_2^2\\
 &\hspace{15pt} \text{subject to} \hspace{10pt} \mathbf{E}{\boldsymbol
 \theta} \leq \mathbf{1}, \hspace{5pt} {\boldsymbol \theta} \geq
 \mathbf{0}, \hspace{5pt} \theta_1 = 1.
\end{split}
\end{equation}
 This is a quadratic programing problem, and efficiently
solved by using any off-the-shelf solver.

For optimizing ${\boldsymbol \alpha}$ with a fixed
${\boldsymbol \theta}^{(t)}$, we solve the following problem:
\begin{align}
\label{eq:SC}
 {\boldsymbol \alpha}^{(t+1)} &= \arg \min_{\boldsymbol \alpha} \|
 \tilde{\mathbf{y}} - {\mathbf{B}} ({\boldsymbol \theta}^{(t)}
 \otimes \mathbf{I}) {\boldsymbol \alpha}\|_2^2 + \eta \|
 {\boldsymbol \alpha }\|_1.
\end{align}
This problem is also an instance of the $\ell^{1}$-norm regularized least
square optimization problem, which is efficiently solved by using sparse coding algorithms. 
We iteratively solve these optimization problems \eqref{eq:QP} and
\eqref{eq:SC} until convergence. 

In Algorithm~\ref{alg:proposed}, we summarize the proposed algorithm
with a pseudo-code. By the operation of {\tt ClipByMatching}, we
estimate the position of the target patch in the auxiliary image
$\mathbf{Y}_{i}$ and extract the auxiliary patch $\mathbf{y}_{i}$ of size
$q^{\prime}$. Also, operations of solving Eqs.~\eqref{eq:QP} and
\eqref{eq:SC} are denoted by {\tt SolveDisp} and {\tt SolveCoeff},
respectively. 
\begin{algorithm}                      
\caption{Proposed Algorithm}
\label{alg:proposed}                     
\begin{algorithmic}
 \State {$\mathbf{Input}$: LR Images
 $\mathbf{Y}_1,\mathbf{Y}_2,\cdots,\mathbf{Y}_N$, HR Dictionary
 $\mathbf{D}^h$, and LR base Dictionaries
 $\mathbf{D}^l,\mathbf{B}$}
 \While{there remains patches to be extracted}
 \State{extract a patch $\mathbf{y}_{1}$ from the target image $\mathbf{Y}_{1}$}
  \For{$i=2$ to $N$}
 \State{${\mathbf{y}_{i}} \leftarrow {\tt
 ClipByMatching}(\mathbf{y}_{1}, \mathbf{Y}_{i})$} 
  \EndFor
   \State{$\tilde{\mathbf{y}} \leftarrow \begin{bmatrix}
										  \tilde{\mathbf{y}},
										  \mathbf{y}_2,
										  \dots,
										  \mathbf{y}_N
 \end{bmatrix}^{\top}$}
   \State{${\boldsymbol \alpha}^{(1)} \leftarrow \arg \min_{\boldsymbol \alpha} \| \mathbf{y}_1 - \mathbf{D}^l {\boldsymbol \alpha}\|_2^2 + \eta \| {\boldsymbol \alpha }\|_1$}
 \For{$t=1$ to $T$}
 \State{${\boldsymbol \theta}^{(t)} \leftarrow {\tt
 SolveDisp}(\tilde{\mathbf{y}}, {\mathbf{B}}, {\boldsymbol
 \alpha}^{(t)}) $}
\Comment{Eqs.~\eqref{eq:QP}}
 \State{${\boldsymbol \alpha}^{(t)} \leftarrow {\tt
 SolveCoeff}(\tilde{\mathbf{y}},{\mathbf{B}}, {\mathbf{\theta}}^{(t)})$}
\Comment{Eqs.~\eqref{eq:SC}}
   \EndFor
   \State{$\mathbf{x} \leftarrow \mathbf{D}^h {\boldsymbol \alpha}^{(T)}$}
 \EndWhile
 \State{post processing for improving the consistency
 of the whole image (see, e.g. \cite{Kato2015})}\\
\Return HR image $\mathbf{X}$
\end{algorithmic}
\end{algorithm}

\subsection{Double sparsity structure}
\begin{figure*}[!th]
 \centering
 \includegraphics[scale=.5]{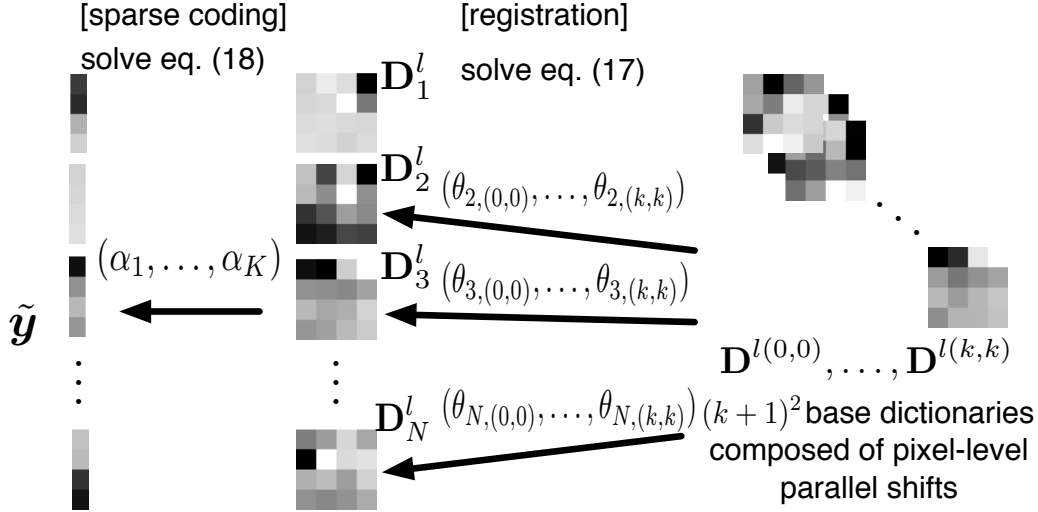}
 \caption{An illustrative diagram for double sparsity for our
  formulation of the multi-frame super resolution problem. The LR
  dictionary used for sparse coding of the stacked LR patches is itself
  a sparse combination of base dictionaries generated by pixel-level
  parallel shift of HR dictionary followed by degradation.
\label{fig:ds}}
 \end{figure*}

It is worth noting that the optimization problem~\eqref{eq:objfunc}
shares the same form with the formulation of {\it{double sparsity}}
dictionary learning
proposed by \cite{rubinstein}. Double sparsity dictionary learning
is proposed for bridging the gap between {\it{analytic}} dictionary such
as wavelets~\cite{Daubechies:1992:TLW:130655} and {\it{learning-based}}
dictionary such MOD~\cite{Engan:1999:MOD:1257298.1257971} and
K-SVD~\cite{ksvd}. The double sparsity
dictionary learning assumes that dictionary atoms themselves have
some underlying sparse structure over a set of more fundamental base dictionaries. In
our formulation of multi-frame super resolution, the atoms are generated
from a sparse combination of fundamental atoms derived by pixel-level shifting and
degenerating the HR atoms. A schematic diagram of the double sparsity
structure in our formulation is shown in Fig.~\ref{fig:ds}. 
By modeling the shift operation by a set of base shift operators, we
obtained a natural and simple formulation of the multi-frame super
resolution based on sparse coding.

\section{Experimental Results}
In this section, we demonstrate the super-resolution results on some sets of still images and some sets of images from movies.

\subsection{Application to still images}
We suppose observed LR images are generated
from an HR image through parallel translations, blurring, down-sampling, and
addition of noises. 
The parallel translations are imitated by shifting
the HR image. The degree of vertical and horizontal shifts are randomly
sampled from a uniform distribution in $[-5,5]$. 
The blurring operation is realized by convolution of $9 \times
9$-pixel Gaussian filter with the standard deviation $\sigma_h = 1$. 
The blurred and shifted images are then down-sampled by the factor
three. Finally, noises sampled from $\mathcal{N}(0,\sigma_{n}=1)$
are added to generate LR observation images.
In our experiments, both the intensity of the blur and
the noise are supposed to be given. 
 In our proposed algorithm, we iteratively solve the
quadratic programming~\eqref{eq:QP} and the sparse coding
problem~\eqref{eq:SC}. We observed that the algorithm converged in less
than three iterations, hence we fix the iteration number $T$ to three in all
of the experiments.
In our experiments, we magnify the input LR images by a factor of three
for all cases. We used five LR images to estimate an HR images, i.e., $N=5$. 

We compare the proposed method to seven conventional methods.
The first method is bi-cubic interpolation. This method is simple
and regarded as a baseline for SR. 
The second and third methods are Single-Frame Joint
Dictionary Learning (SF-JDL;~\cite{Yang:2010:ISV:1892456.1892463}) and Adaptive Sparse Domain
Selection (ASDS;~\cite{asdssr}). These methods are considered as state-of-the-art
single-frame SR methods with publicly available software implementations.
The fourth method is the one proposed
by~\cite{wang1}, which is a
multi-frame SR method based on joint dictionary learning. We refer to
this method as MF-JDL (Multi-Frame super resolution based on Joint
Dictionary Learning). The other two methods are representative methods
in reconstruction-based SR in the literature. In~\cite{frsr}, a multi-frame SR method
based on regularization in the form of Bilateral Total Variation (BTV) is
proposed. Because of its performance and simplicity, the method have
become one of the most commonly cited papers in the field of multi-frame
SR. BTV is further improved in~\cite{Li:2010:MIS:1621135.1621163}, in which the method based on
regularization by Locally Adaptive Bilateral Total Variation (LABTV) is
proposed. In this paper, these two reconstruction-based methods are referred to as BTV
and LABTV, respectively. Finally, 
we also use the multi-frame SR method based on sparse coding proposed in our
previous paper~\cite{Kato2015}, which is referred to as MF-SC.

There are several tuning parameters for each SR methods. To make fair
comparison, we first optimize the parameters of each method to maximize
PSNR of the image Lena, which is one of the most commonly used
benchmark images in the field of image processing. Then, for all other images, we keep using
the same parameters which are optimized for Lena.

We use two different gray-scale images (Lena and Cameraman), and three
color images (Flower, Girl, and Parthenon) for evaluating the
performance of SR methods. 
 When we deal with color images, we first
convert the image into YCbCr format, then apply SR methods only to
luminance channel (Y). Values of other channels Cb and Cr are simply
expanded by bi-cubic interpolation. 

We show the experimental results in
Fig.~\ref{r_lena}-Fig.~\ref{r_parthenon}. To focus on the difference between our previous method and the newly
proposed method, we only show the original images, degraded images,
images obtained by MF-SC and those obtained by our proposed method. 
 These figures indicate that the proposed method can generate
 comparable or better images compared to MF-SC.

{\setlength{\tabcolsep}{1mm}
\begin{figure}[!ht]
  \begin{center}
    \begin{tabular}{cc}
      \begin{minipage}{4cm}
        \begin{center}
          \includegraphics[clip, width = 4cm]{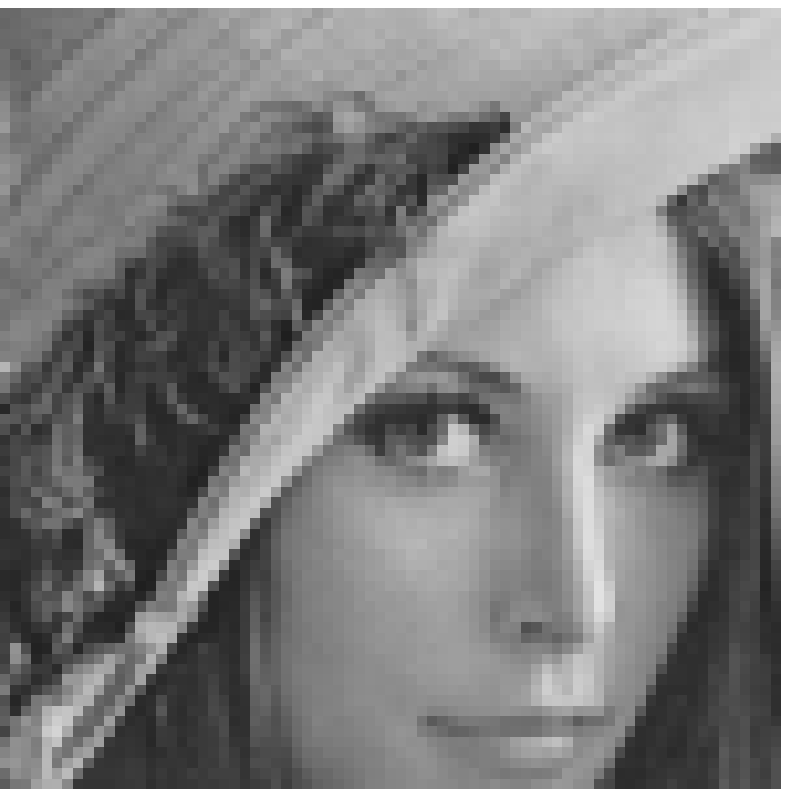}
        \end{center}
      \end{minipage}
	&
      \begin{minipage}{4cm}
        \begin{center}
          \includegraphics[clip, width = 4cm]{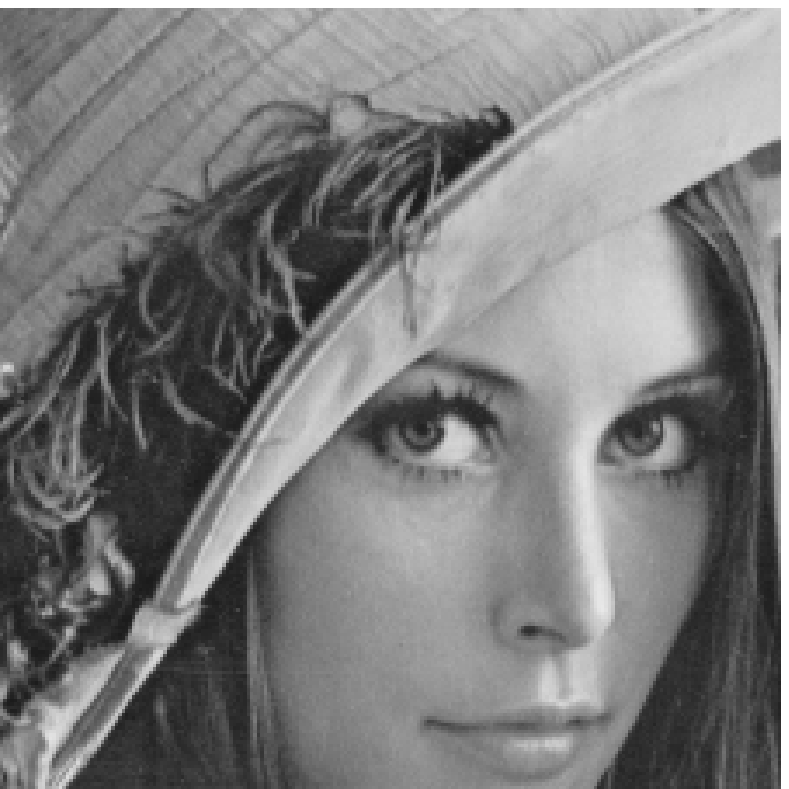}
        \end{center}
      \end{minipage}
\\
	{\small (a) Observed LR image}
	&
	{\small (b) Original HR image}
		 \\
  \begin{minipage}{4cm}
        \begin{center}
          \includegraphics[clip, width = 4cm]{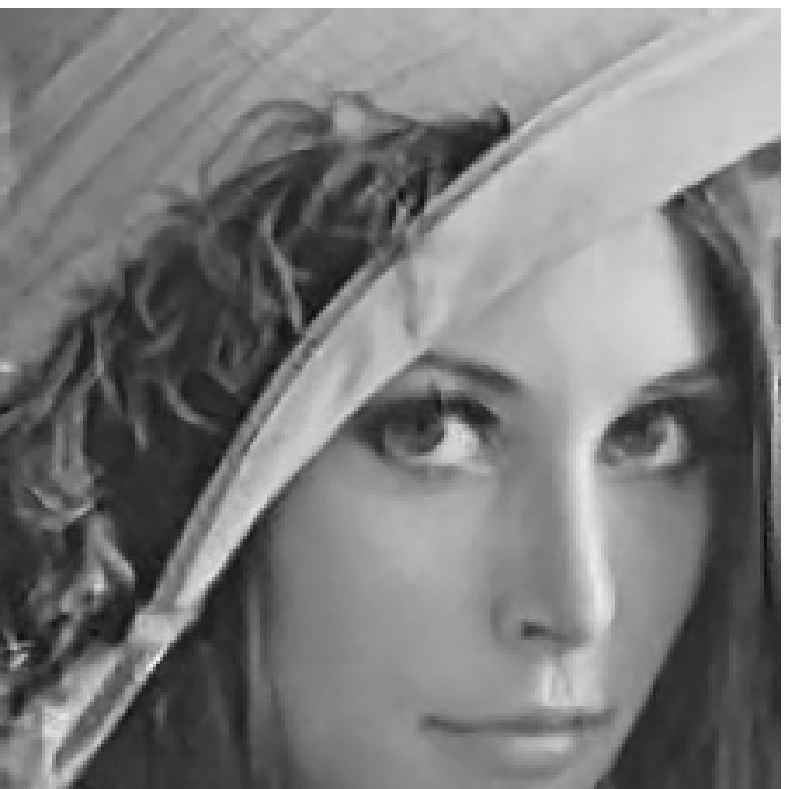}
        \end{center}
      \end{minipage}
	&
      \begin{minipage}{4cm}
        \begin{center}
          \includegraphics[clip, width = 4cm]{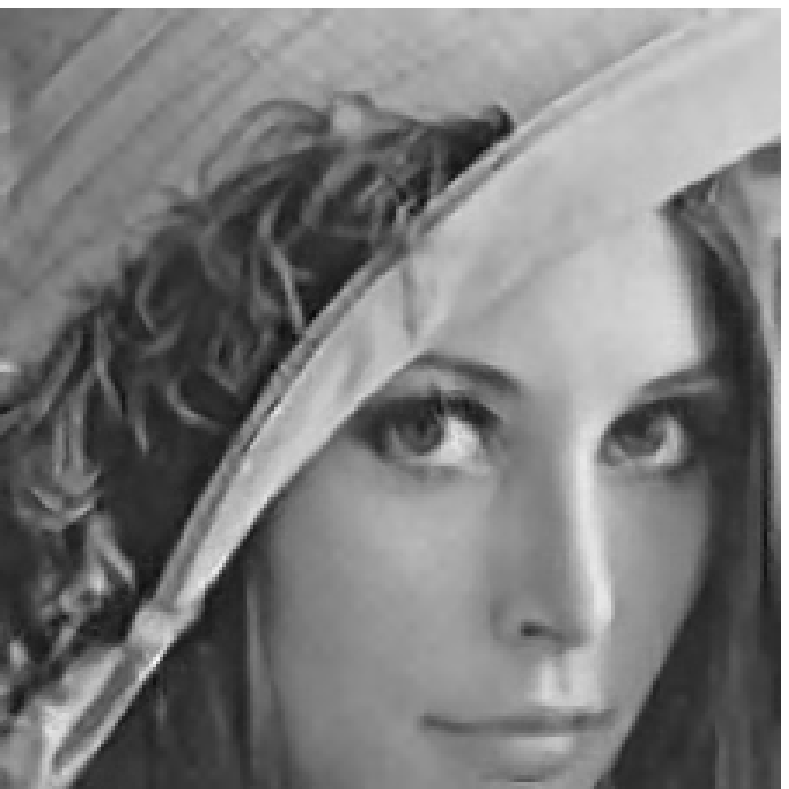}
        \end{center}
      \end{minipage}
		\\

	{\small (c) MF-SC}
	&
	{\small (d) Proposed}
    \end{tabular}
  \caption{Reconstructed images estimated from LR observations for
   Lena. 
(a) Observed LR image, (b) Original HR image, results by (c) MF-SC(our
   previous work), and (d) the proposed method.
\label{r_lena}}
  \end{center}
\end{figure}
}

{\setlength{\tabcolsep}{1mm}
\begin{figure}[!ht]
  \begin{center}
    \begin{tabular}{cc}
      \begin{minipage}{4cm}
        \begin{center}
          \includegraphics[clip, width = 4cm]{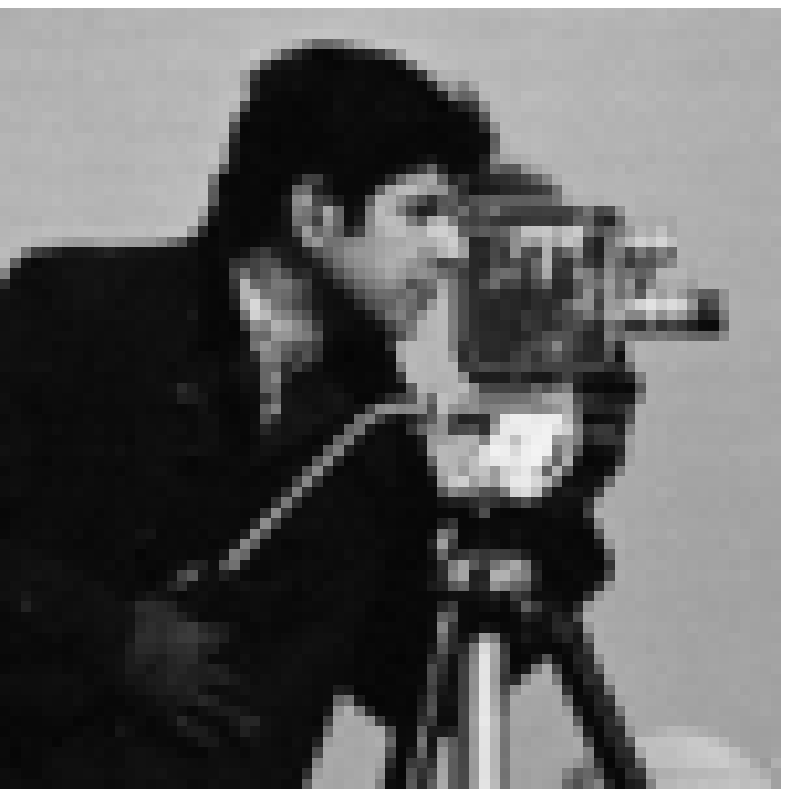}
        \end{center}
      \end{minipage}
	&
      \begin{minipage}{4cm}
        \begin{center}
          \includegraphics[clip, width = 4cm]{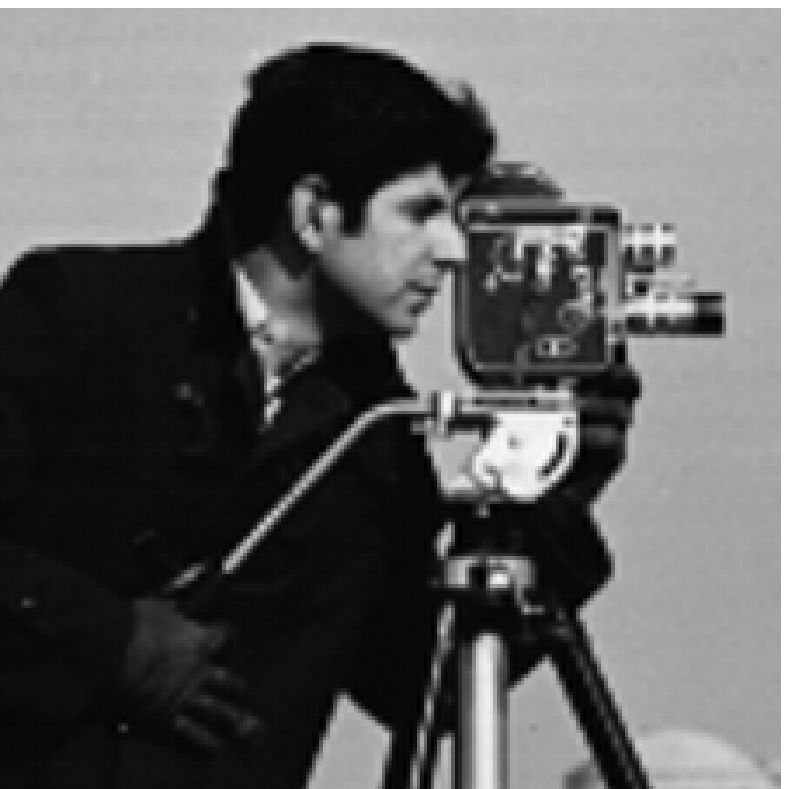}
        \end{center}
      \end{minipage}
\\
	{\small (a) Observed LR image}
	&
	{\small (b) Original HR image}
		 \\
      \begin{minipage}{4cm}
        \begin{center}
          \includegraphics[clip, width = 4cm]{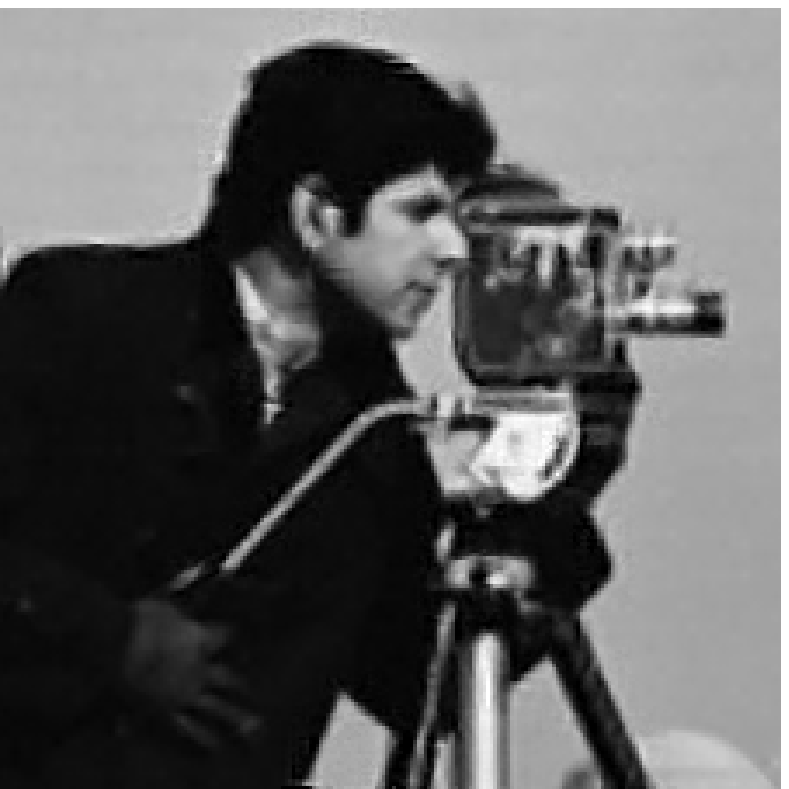}
        \end{center}
      \end{minipage}
	&
      \begin{minipage}{4cm}
        \begin{center}
          \includegraphics[clip, width = 4cm]{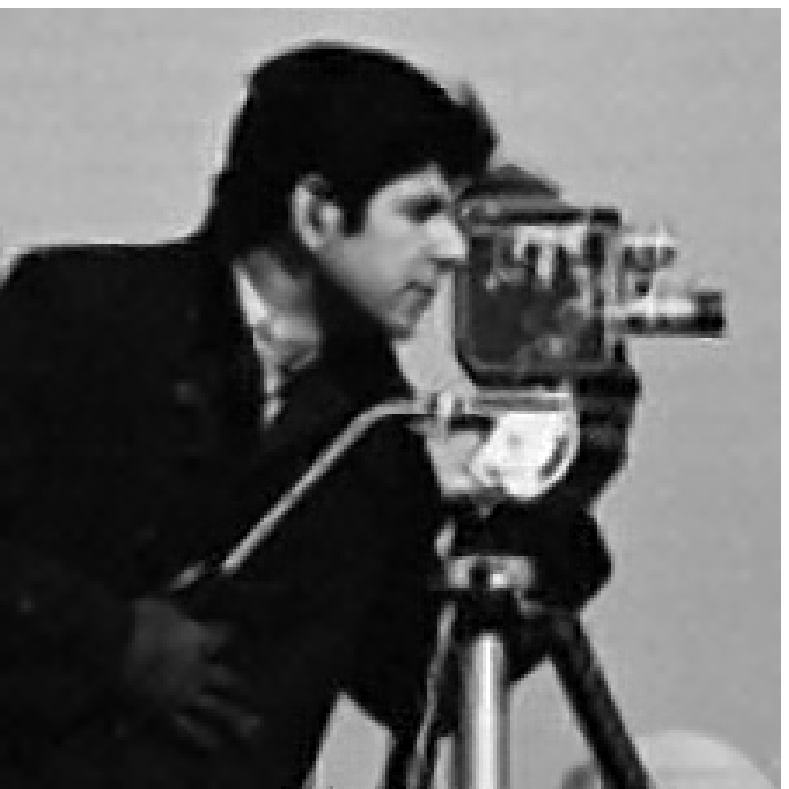}
        \end{center}
      \end{minipage}
\\
	{\small (c) MF-SC}
	&
	{\small (d) Proposed}
    \end{tabular}
 \caption{Reconstructed images estimated from LR observations for
   Cameraman. 
(a) Observed LR image, (b) Original HR image, results by (c) MF-SC, and (d) the proposed method. \label{r_cameraman}}
  \end{center}
\end{figure}
}

{\setlength{\intextsep}{0.5pt}
{\setlength{\tabcolsep}{1mm}
\begin{figure}[!h]
  \begin{center}
    \begin{tabular}{cc}

      \begin{minipage}{4cm}
        \begin{center}
          \includegraphics[clip, width = 4cm]{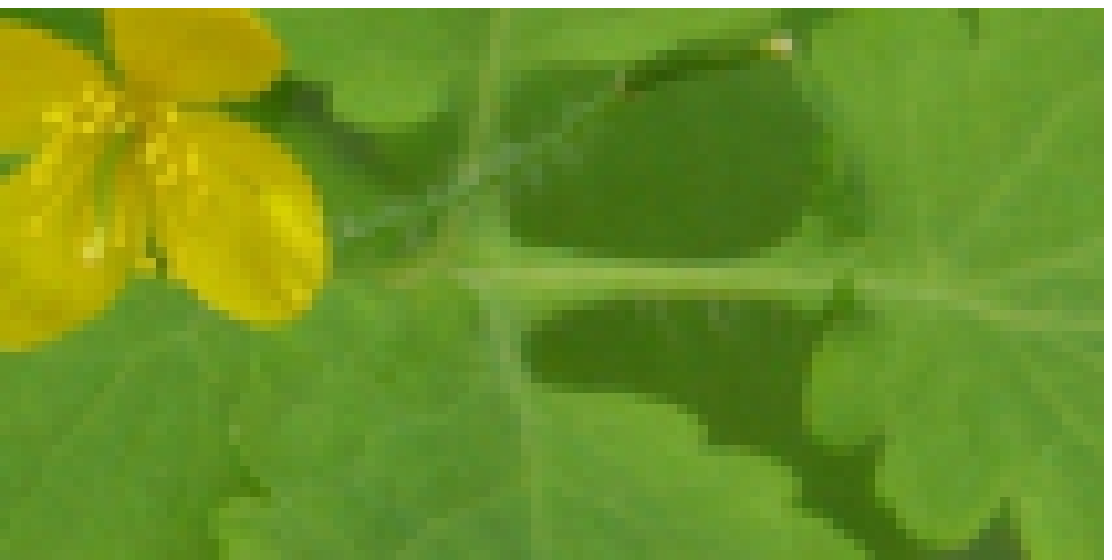}
        \end{center}
      \end{minipage}

	&
      \begin{minipage}{4cm}
        \begin{center}
          \includegraphics[clip, width = 4cm]{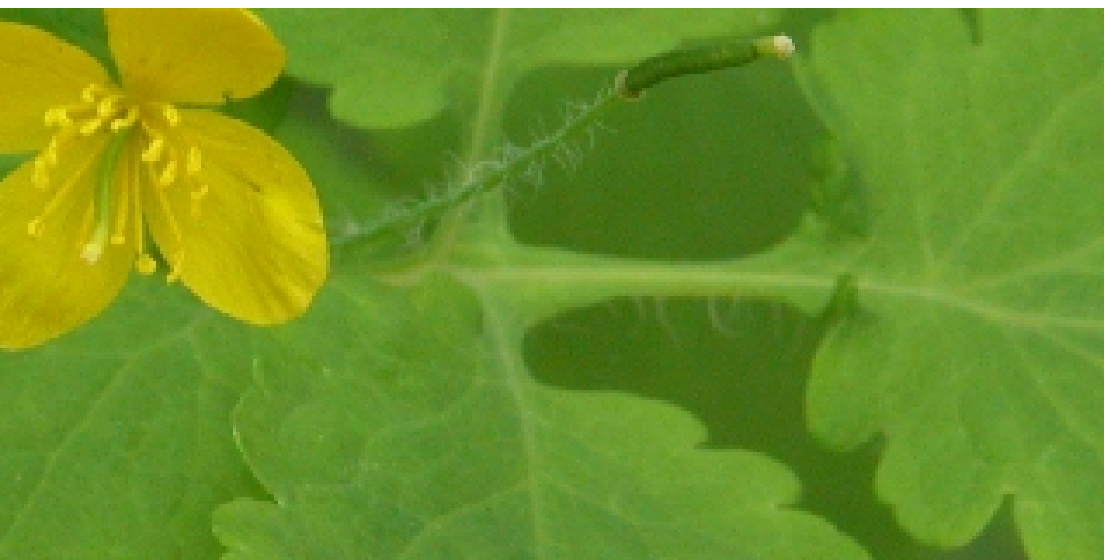}
        \end{center}
      \end{minipage}
\\

	{\small (a) Observed LR image}
	&
	{\small (b) Original HR image}
		\\
      \begin{minipage}{4cm}
        \begin{center}
          \includegraphics[clip, width = 4cm]{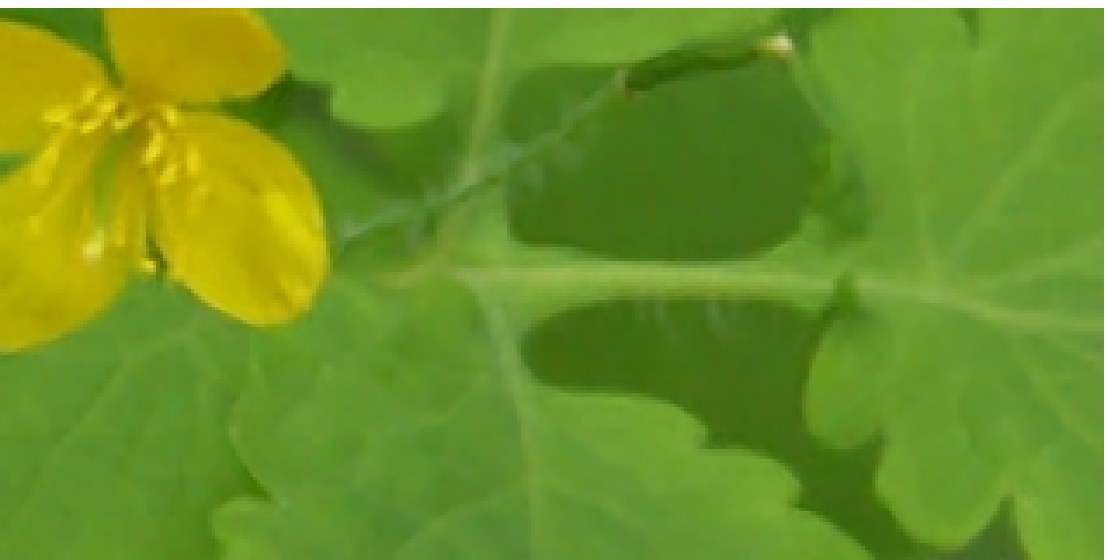}
        \end{center}
      \end{minipage}

	&
      \begin{minipage}{4cm}
        \begin{center}
          \includegraphics[clip, width = 4cm]{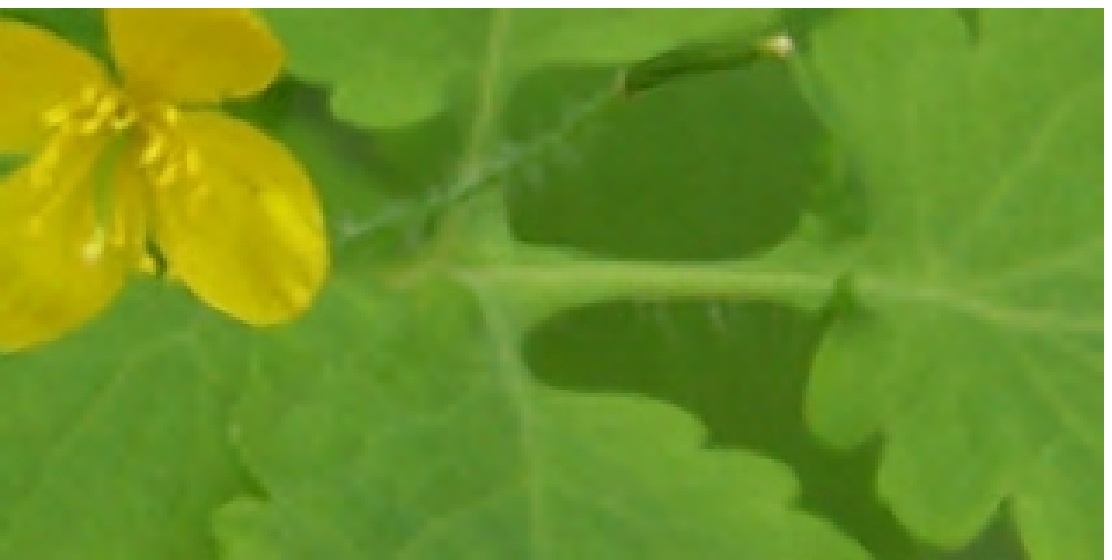}
        \end{center}
      \end{minipage}   
\\
	{\small (c) MF-SC}
	&
	{\small (d) Proposed}
    \end{tabular}
\caption{Reconstructed images estimated from LR observations for
   Flower. 
(a) Observed LR image, (b) Original HR image, results by (c) MF-SC, and (d) the proposed method.
\label{r_flower}}
  \end{center}
\end{figure}
}
}

{\setlength{\intextsep}{0.5pt}
{\setlength{\tabcolsep}{1mm}
\begin{figure}[!h]
  \begin{center}
    \begin{tabular}{cc}
      \begin{minipage}{4cm}
        \begin{center}
          \includegraphics[clip, width = 4cm]{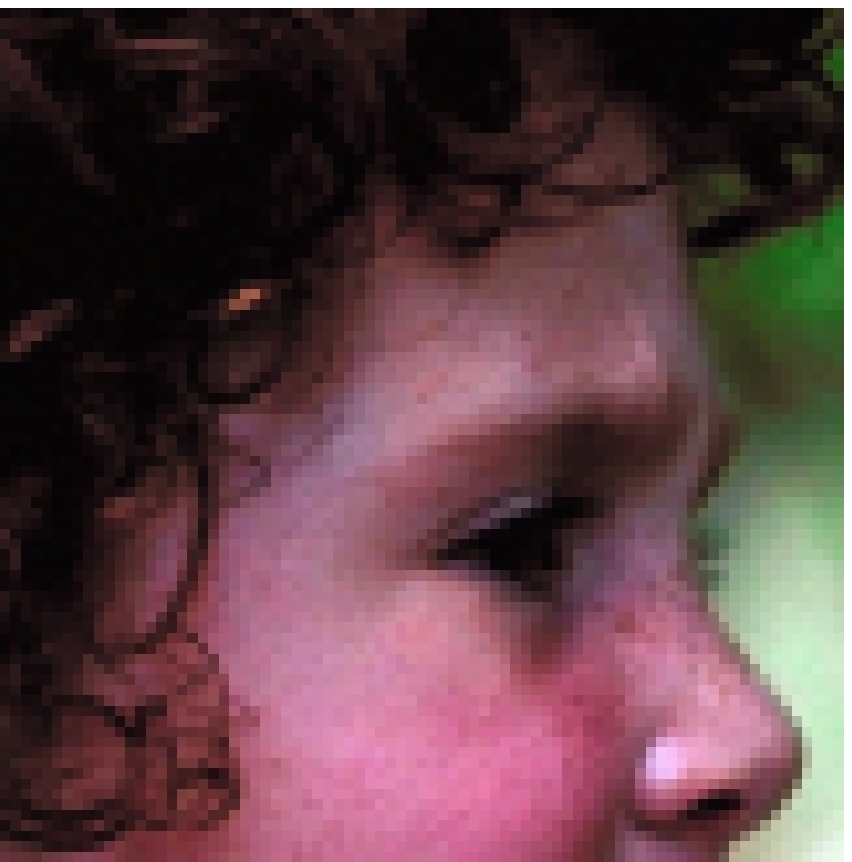}
        \end{center}
      \end{minipage}
	&
      \begin{minipage}{4cm}
        \begin{center}
          \includegraphics[clip, width = 4cm]{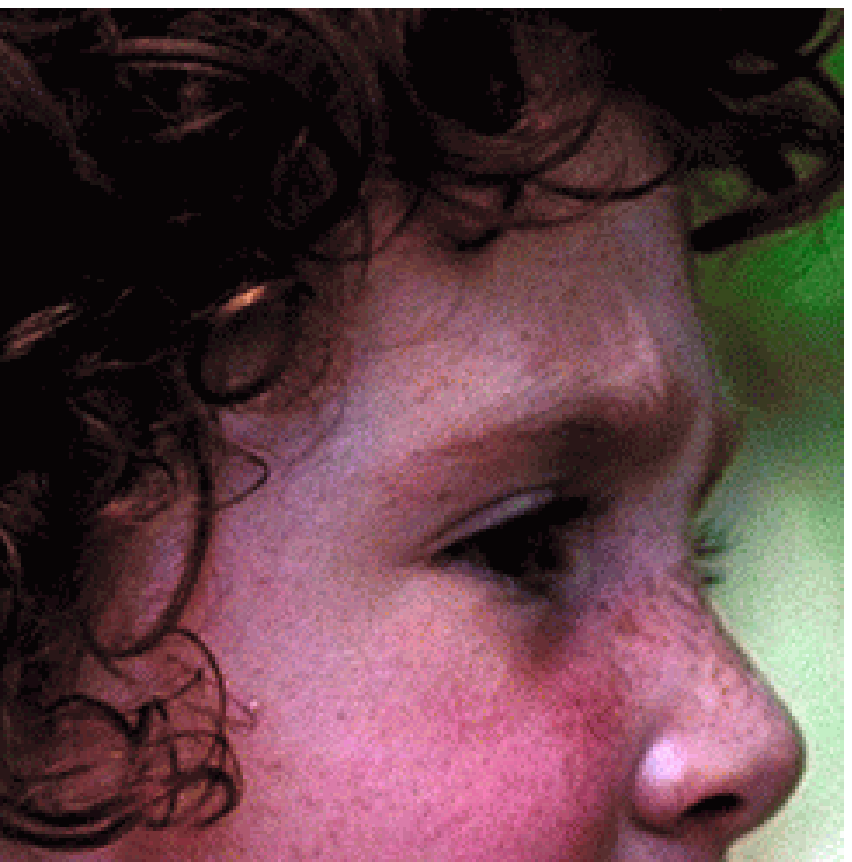}
        \end{center}
      \end{minipage}
\\
	{\small (a) Observed LR image}
	&
	{\small (b) Original HR image}
\\
     \begin{minipage}{4cm}
        \begin{center}
          \includegraphics[clip, width = 4cm]{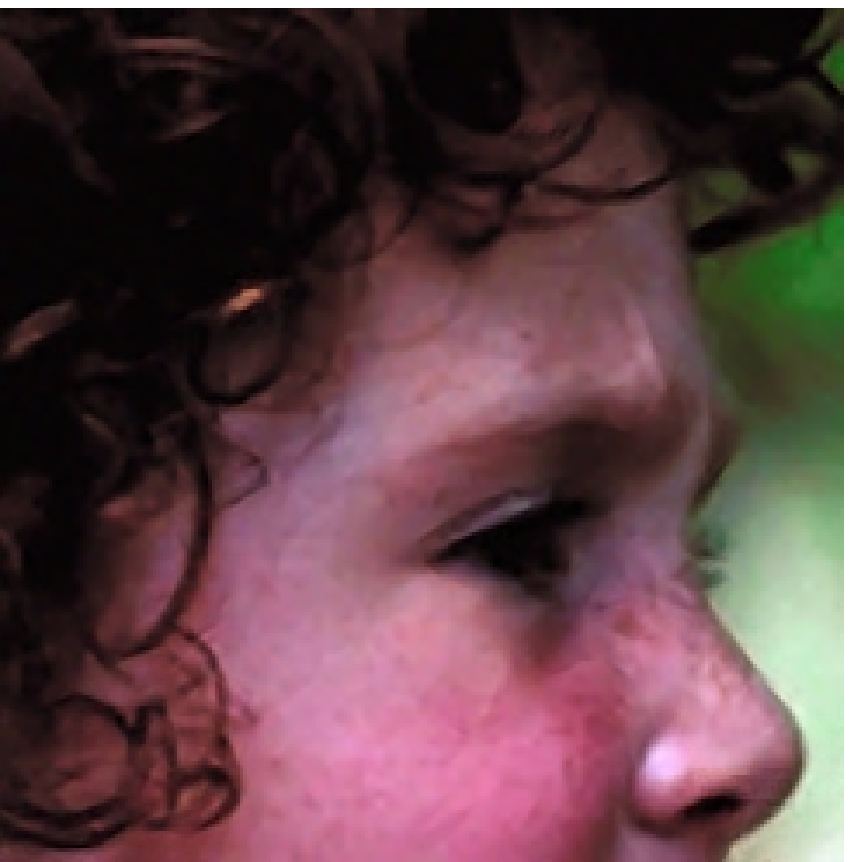}
        \end{center}
      \end{minipage}
	&
      \begin{minipage}{4cm}
        \begin{center}
          \includegraphics[clip, width = 4cm]{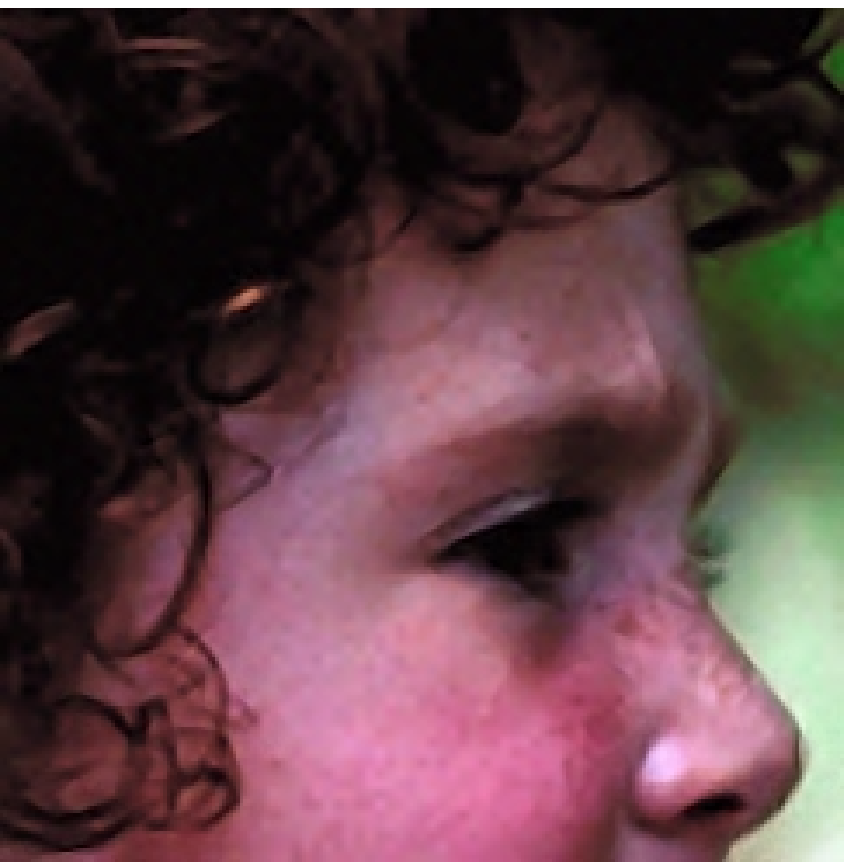}
        \end{center}
      \end{minipage}
		\\

	{\small (c) MF-SC}
	&
	{\small (d) Proposed}
    \end{tabular}
\caption{Reconstructed images estimated from LR observations for
   Girl. 
(a) Observed LR image, (b) Original HR image, results by (c) MF-SC, and (d) the proposed method.
\label{r_girl}}
  \end{center}
\end{figure}
}
}

{\setlength{\intextsep}{0.5pt}
{\setlength{\tabcolsep}{1mm}
\begin{figure}[!h]
  \begin{center}
    \begin{tabular}{cc}

      \begin{minipage}{4cm}
        \begin{center}
          \includegraphics[clip, width = 4cm]{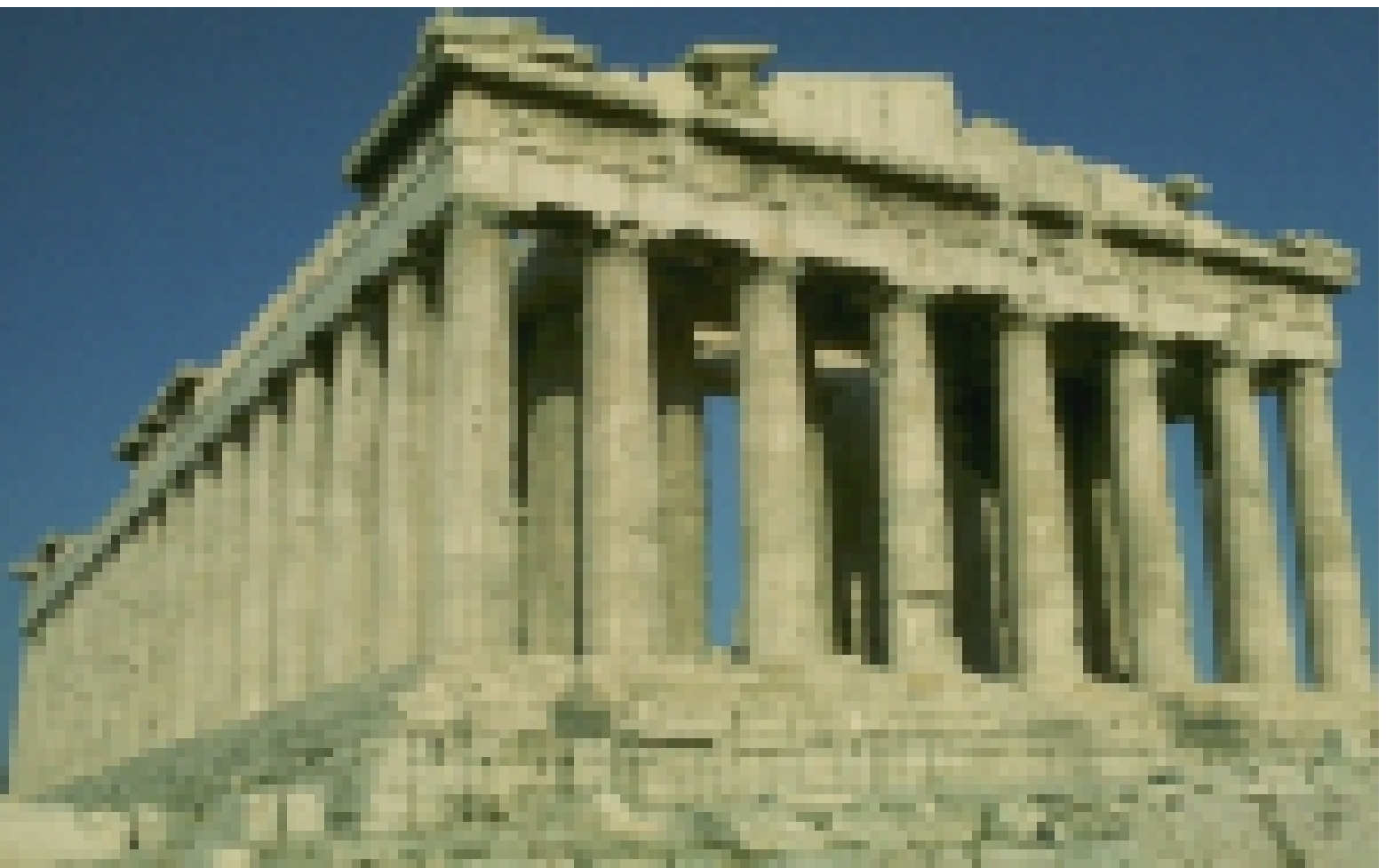}
        \end{center}
      \end{minipage}

	&
      \begin{minipage}{4cm}
        \begin{center}
          \includegraphics[clip, width = 4cm]{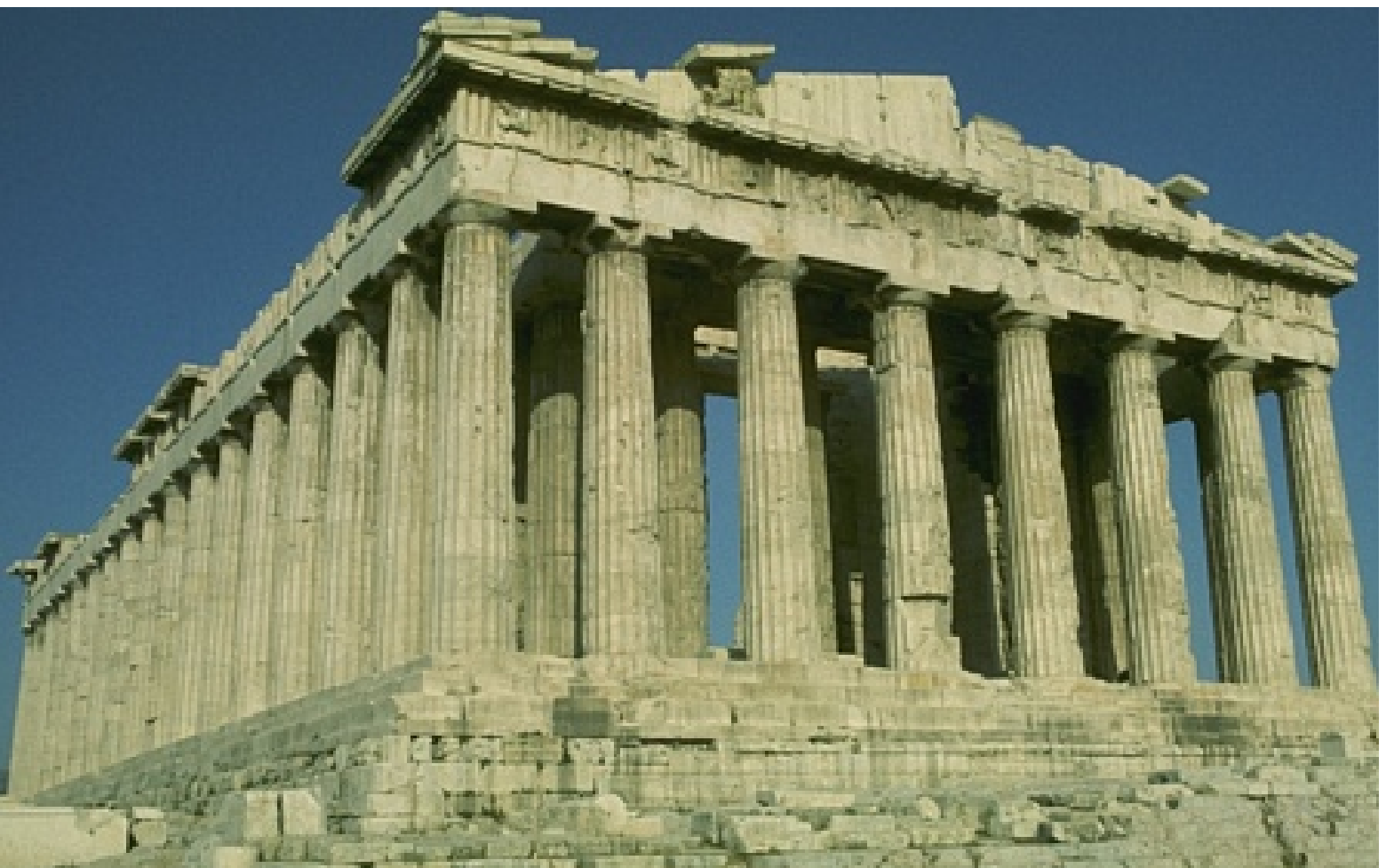}
        \end{center}
      \end{minipage}
\\
	{\small (a) Observed LR image}
	&
	{\small (b) Original HR image}
\\
           \begin{minipage}{4cm}
        \begin{center}
          \includegraphics[clip, width = 4cm]{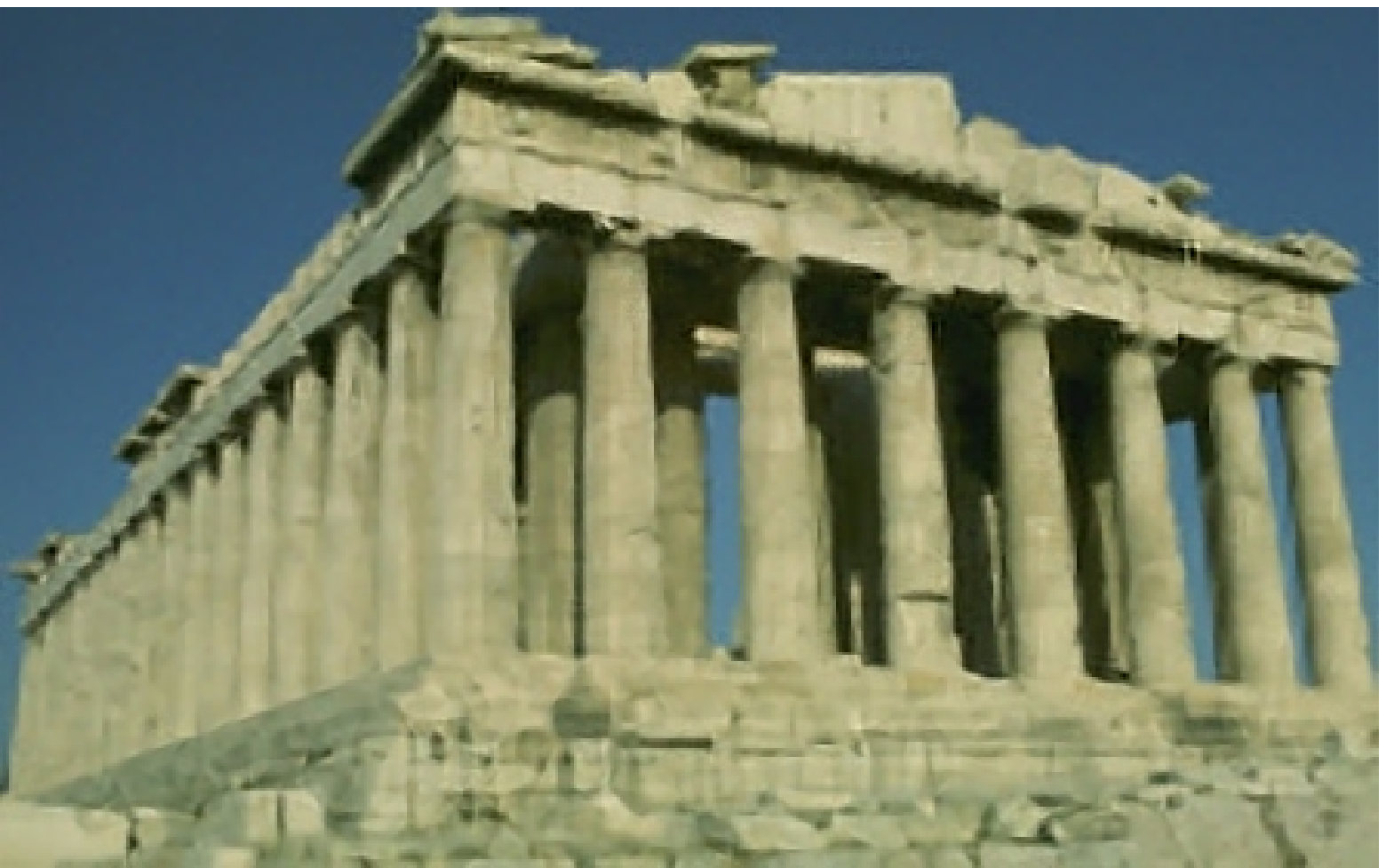}
        \end{center}
      \end{minipage}

	&
      \begin{minipage}{4cm}
        \begin{center}
          \includegraphics[clip, width = 4cm]{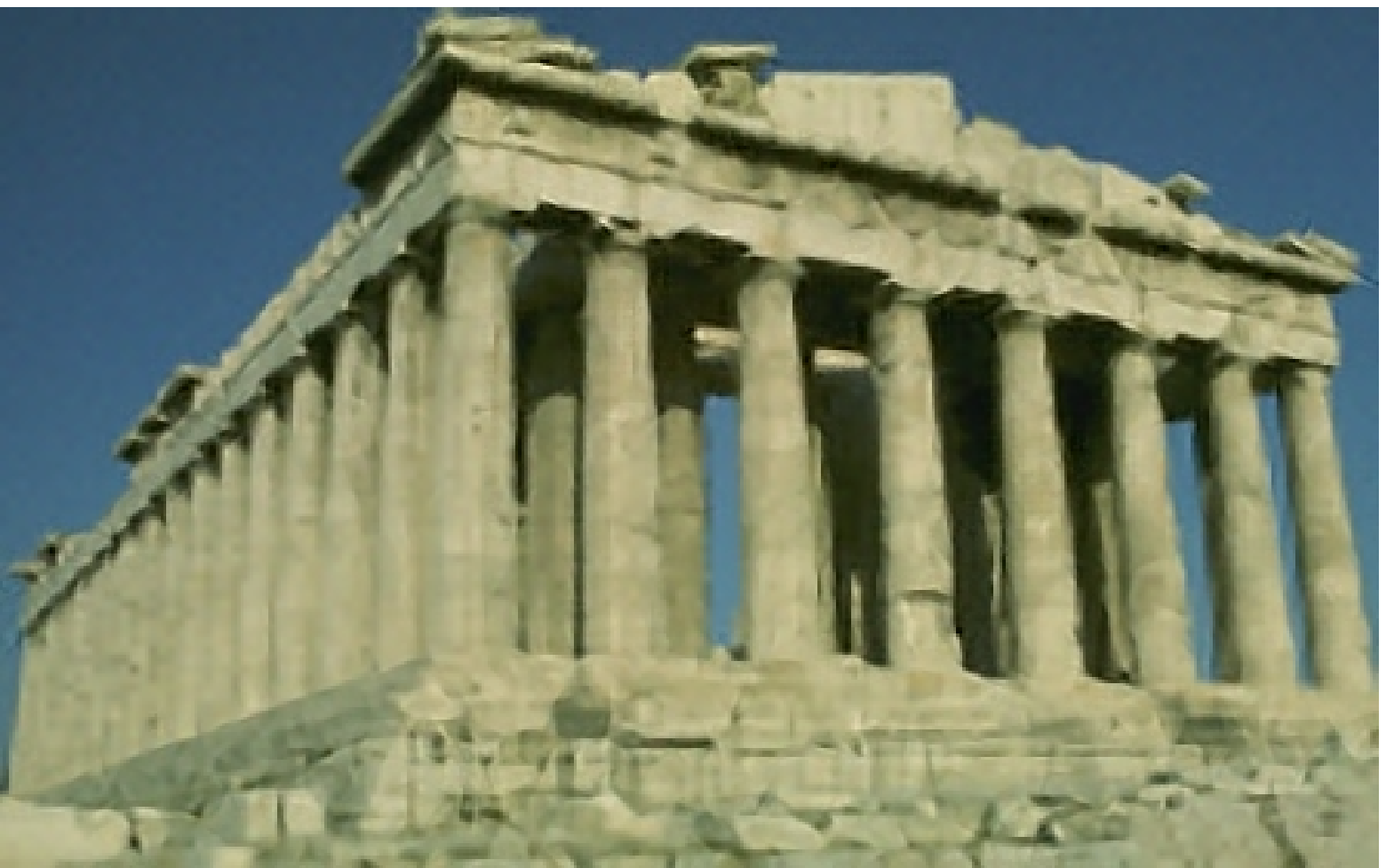}
        \end{center}
      \end{minipage}

	\\
	{\small (c) MF-SC}
	&
	{\small (d) Proposed}
    \end{tabular}
\caption{Reconstructed images estimated from LR observations for
   Parthenon.
(a) Observed LR image, (b) Original HR image, results by (c) MF-SC, and (d) the proposed method.
\label{r_parthenon}}
  \end{center}
\end{figure}
}
}

For quantitative comparison of SR methods, we use the Peak Signal to
Noise Ratio (PSNR) defined as
\begin{align}
{\rm{PSNR}}\text{[dB]} = 10 \log_{10} \frac{255^2}{{\rm{MSE}}},
\end{align}
where MSE is the mean squared error between the original HR image and
the estimated HR image, and the higher PSNR indicates the better SR
performance.
We show PSNR values obtained by various methods in Table~\ref{tab:result_psnr}.
For evaluating the PSNR values, we randomly generated $100$ sets of $5$
shift operators and generated 
degraded images by adding random observation noises. From each set of observed images, we randomly choose one target image. 
The only target image is used for single-frame SR, while in multi-frame SR, the remaining $4$ images are used as auxiliary LR images.
The means and standard deviations of PSNR values are calculated using
$100$ SR results by each methods. The best and the second best results
are shown in bold and underlined styles, respectively.
\begin{table*}[th!]
\begin{center}
   \caption{PSNRs of SR method. 
The best and second best results are shown by bold and underlined
 styles, respectively.\label{tab:result_psnr}}
\scalebox{.7}{
\begin{tabular}{c|cccccccc} \hline
	Image & Bicubic & SF-JDL & ASDS & MF-JDL & BTV & LABTV & MF-SC& Proposed \\ \hline
	Lena & $27.91\pm 0.00$ & $28.73 \pm 0.01$ & $\bold{30.08} \pm 0.02$
			 & $29.25 \pm 0.05$ & $29.01 \pm 0.22$ & $29.33 \pm 0.20$ &
							 $ 29.69 \pm 0.15$ & $\underline{29.82} \pm 0.14$\\
	Cameraman & $27.03 \pm 0.00$ & $28.25 \pm 0.01$ & $29.88
			 \pm 0.02$ & $28.29 \pm 0.03$ & $29.43 \pm 0.37$ & $29.83
						 \pm 0.37$ & $\underline{30.19} \pm 0.38$ & ${\bold{30.44}} \pm 0.35$\\
	Flower & $35.50 \pm 0.01$ & $35.81 \pm 0.01$ & $36.22 \pm 0.02$ &
				 $36.32 \pm 0.04$ & $36.26 \pm 0.24$ &
						 $\underline{36.46} \pm 0.21$ & $\bold{36.61}
							 \pm 0.10$ &
 $\bold{36.61} \pm 0.12$\\
	Girl & $31.12 \pm 0.00$ & $31.49 \pm 0.01$ & $31.72 \pm 0.01$ &
				 $31.73 \pm 0.02$ & $31.84 \pm 0.17$ & $\bold{32.09} \pm
						 0.16$ & $\underline{31.98} \pm 0.06$ 
& $31.92 \pm 0.07$\\
	Parthenon & $24.40 \pm 0.00$ & $24.59 \pm 0.00$ & $25.07 \pm 0.01$&
				 $24.70 \pm 0.02$ & $\bold{25.45} \pm 0.19$ & $25.33 \pm
						 0.13$ & $\underline{25.41} \pm 0.09$ 
& $25.37 \pm 0.08$\\ \hline
 \end{tabular}
}
\end{center}
\end{table*}
As shown in Table~\ref{tab:result_psnr}, the proposed method 
outperforms other conventional methods in two out of five images (
Cameraman and Flower), and
be the second best for Lena. It improved the previous method in two
images, being the same in one image, and slightly worth in two images.

\subsection{Application to motion pictures}
We show experimental results on LR images sequentially captured from movies. 
From five consecutive LR images, the middle (third in the temporal
sequence) image is selected as the target image, and other four are
considered as auxiliary images.

\begin{figure}[!htb]
  \begin{center}
    \begin{tabular}{cc}
     \begin{minipage}{4.0cm}
        \begin{center}
          \includegraphics[clip, width=4.0cm]{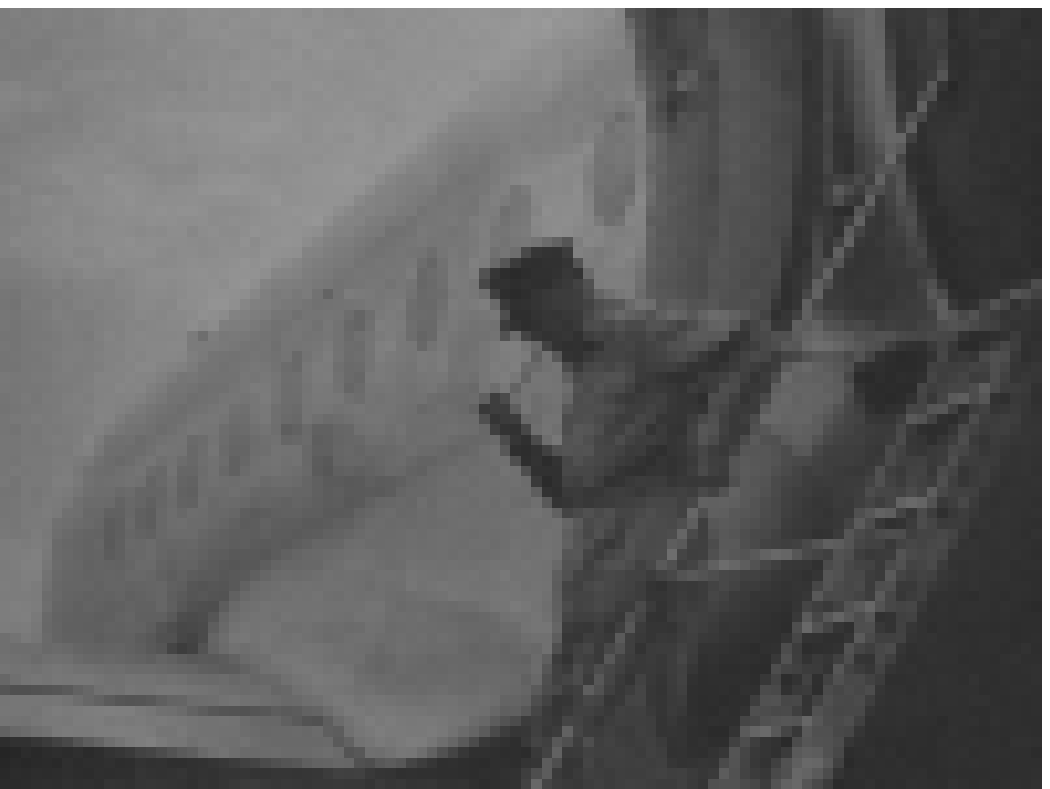}
        \end{center}
      \end{minipage}
&
     \begin{minipage}{4.0cm}
        \begin{center}
          \includegraphics[clip, width=4.0cm]{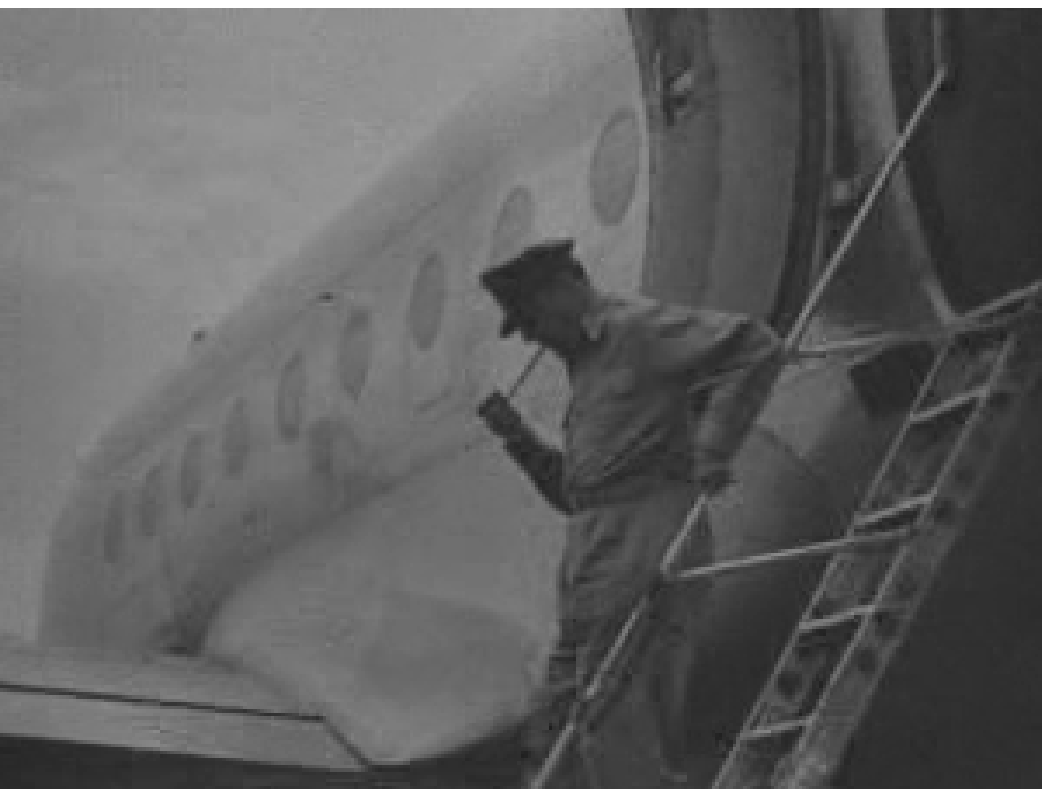}
        \end{center}
      \end{minipage}
		\\
	{\small (a) Observed LR image}
	&
	{\small (b) Original HR image}
\\
     \begin{minipage}{4.0cm}
        \begin{center}
          \includegraphics[clip, width=4.0cm]{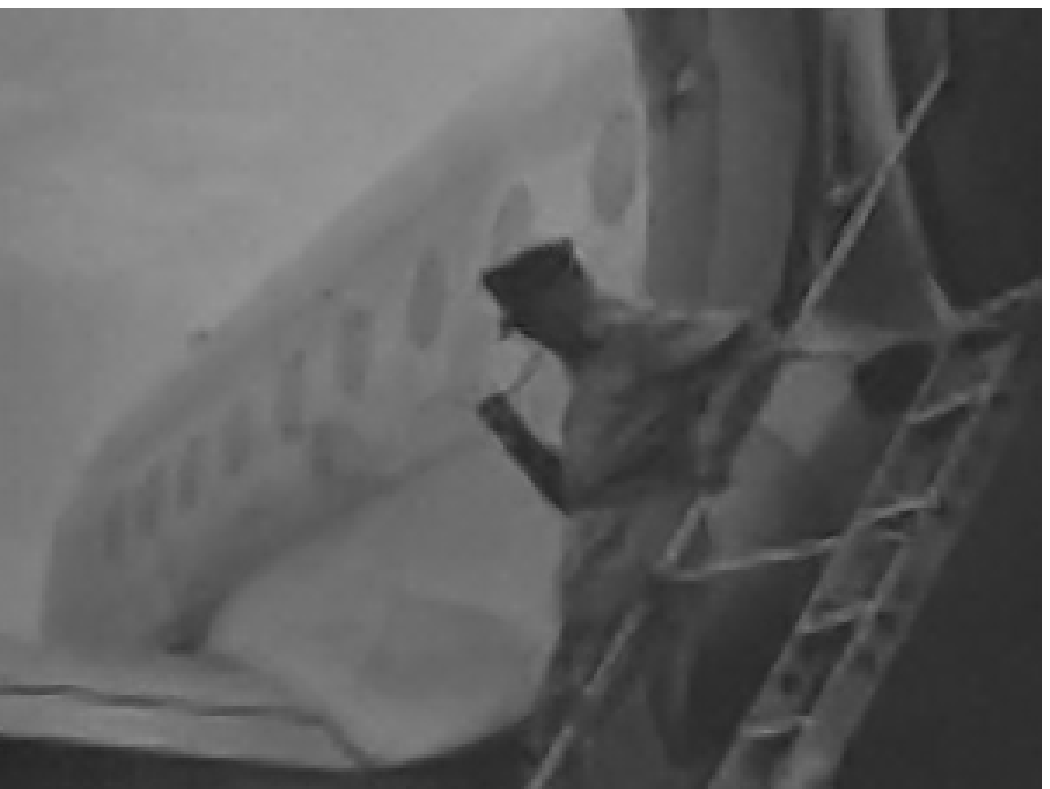}
        \end{center}
      \end{minipage}
&
     \begin{minipage}{4.0cm}
        \begin{center}
          \includegraphics[clip, width=4.0cm]{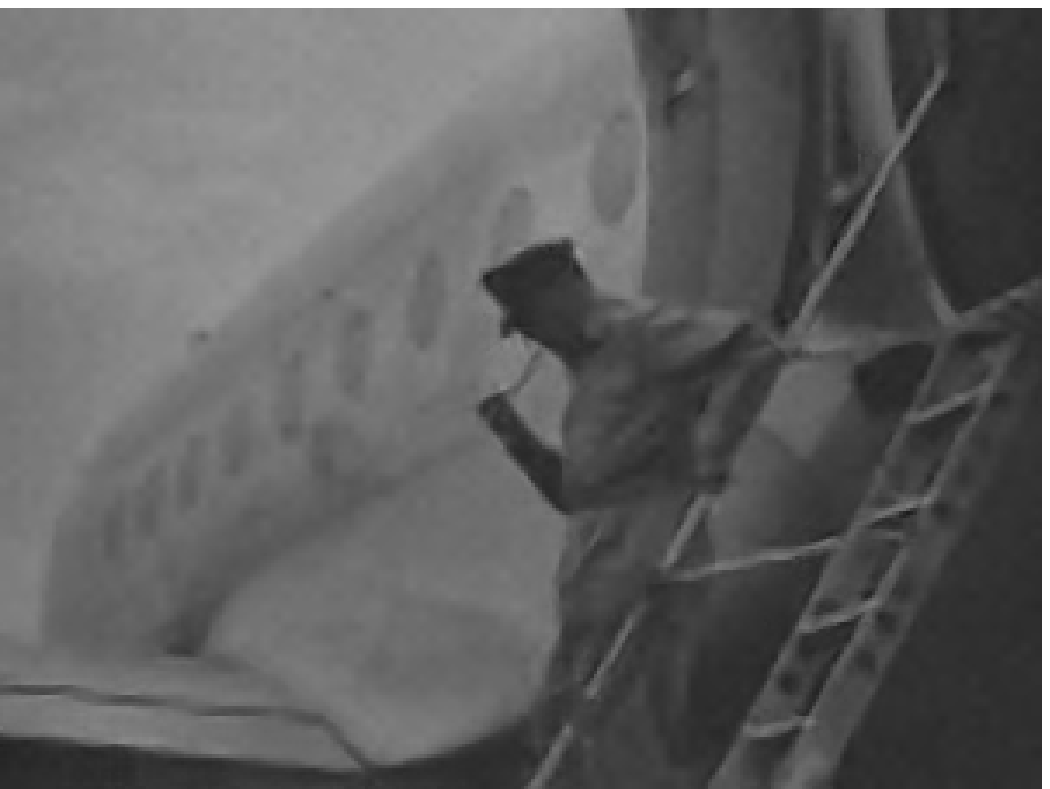}
        \end{center}
      \end{minipage}

\\
	{\small (c) MF-SC}
&
	{\small (d) Proposed}
    \end{tabular}
\caption{Images estimated from LR observations for MacArthur.
(a) Observed LR image, (b) Original HR image, results by (c) MF-SC, and (d) the proposed method.
\label{r_mac}}
  \end{center}
\end{figure}

\begin{figure}[!htb]
  \begin{center}
    \begin{tabular}{cc}
      \begin{minipage}{4.0cm}
        \begin{center}
          \includegraphics[clip, width=4.0cm]{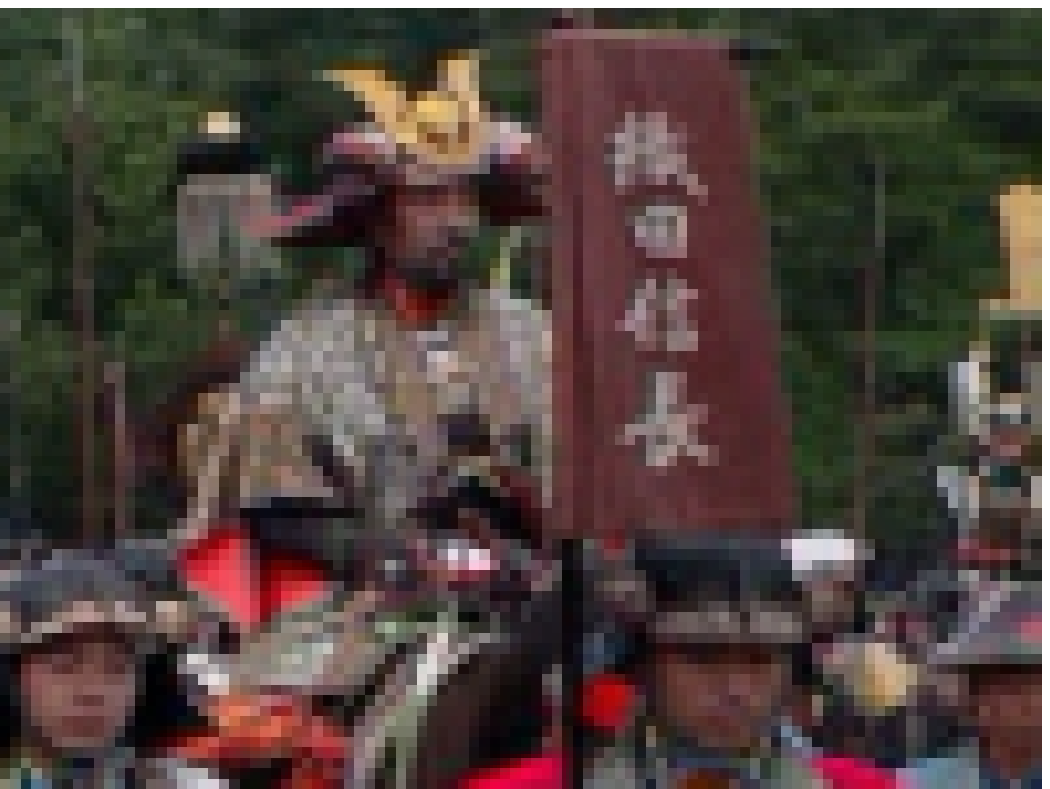}
        \end{center}
      \end{minipage}
&
      \begin{minipage}{4.0cm}
        \begin{center}
          \includegraphics[clip, width=4.0cm]{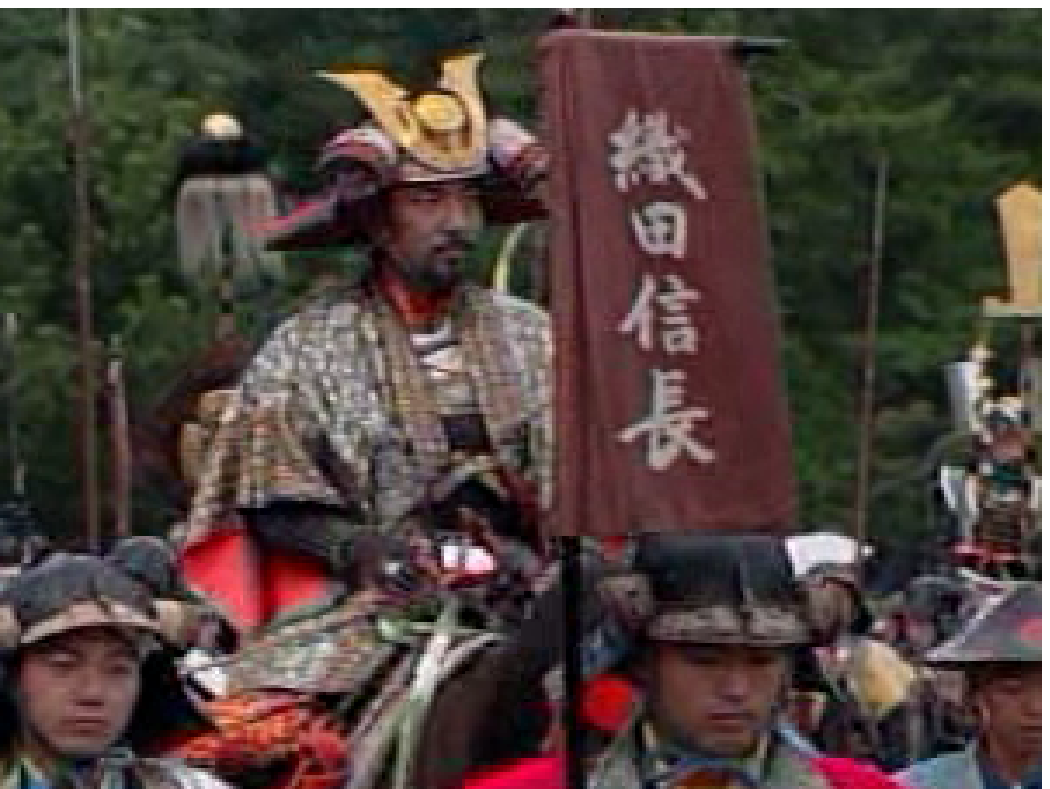}
        \end{center}
      \end{minipage}

\\
	{\small (a) Observed LR image}
&
	{\small (b) Original HR image}
\\
      \begin{minipage}{4.0cm}
        \begin{center}
          \includegraphics[clip, width=4.0cm]{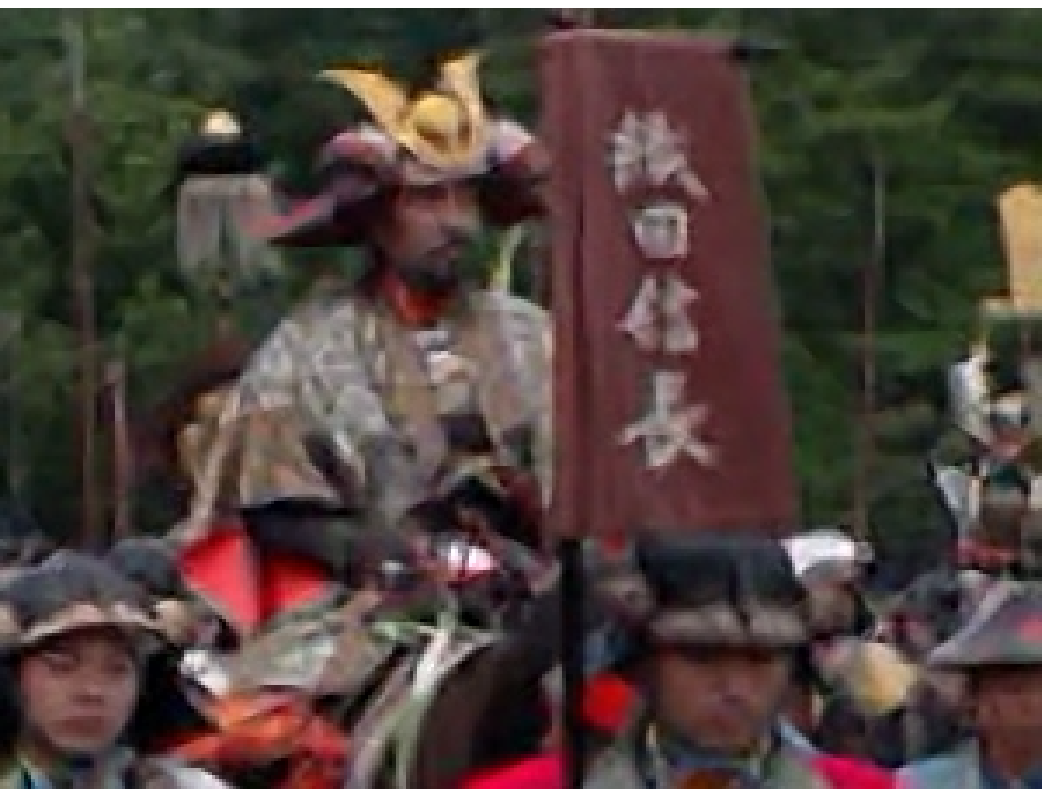}
        \end{center}
      \end{minipage}
&
      \begin{minipage}{4.0cm}
        \begin{center}
          \includegraphics[clip, width=4.0cm]{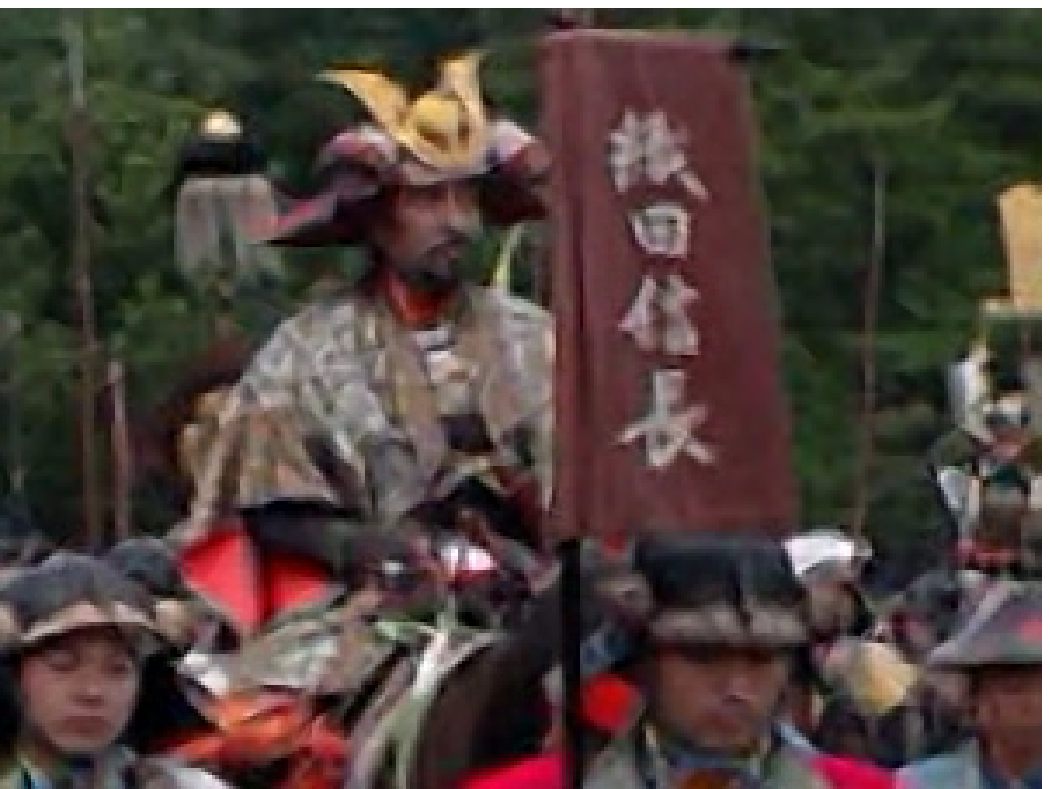}
        \end{center}
      \end{minipage}

\\
	{\small (c) MF-SC}
&
	{\small (d) Proposed}
    \end{tabular}
\caption{Images estimated from LR observations for Samurai.
(a) Observed LR image, (b) Original HR image, results by (c) MF-SC, and (d) the proposed method.
\label{r_samurai}}
  \end{center}
\end{figure}

 The obtained HR images using MF-SC and the proposed method are shown in 
Fig.~\ref{r_mac} and Fig.~\ref{r_samurai}.
We also show the obtained PSNR in Table~\ref{tab:psnr_movie} for all methods.

\begin{table*}[!th]
\begin{center}
	\caption{PSNRs and computational times (in parentheses) of SR methods applied to movie data.}
 \label{tab:psnr_movie}
\begin{tabular}{c|ccccccccc} \hline
 & Bicubic & SF-JDL & ASDS & MF-JDL & BTV & LABTV & MF-SC &Proposed \\ \hline
 MacArthur & 34.11 & 34.33  & 35.63   & 35.18 &
					 34.39  & 34.40  & 34.79  & 35.63\\
 & & (2.69) & (178.08)  & (133.78) & (61.72) & (96.17) &  (27.70) & (61.74)\\
 
 Samurai & 25.36 & 25.97 & 26.66  & 26.12 & 26.16
					  & 26.07 & 25.90 & 26.49\\ 
  &  & (2.50) & (211.65)  & (138.38) &  (62.13) &  (96.24) &  (30.75) &(59.86)\\ \hline
 \end{tabular}
 \end{center}
\end{table*}

From Fig.~\ref{r_mac} and Fig.~\ref{r_samurai}, the HR images obtained by the proposed method
are clear and have distinct edges compared to the images obtained by
MF-SC.
From Table~\ref{tab:psnr_movie}, PSNRs of the proposed method are the
same or lower
than ASDS. However, the computational costs of the
proposed method are lower than ASDS and other multi-frame
SR methods.
Although the proposed method requires about double computational cost to our
previous method (MF-SC), it significantly improves the quality of
reconstructed image in PSNR.

\section{Conclusion}
In this paper, we discussed multi-frame image super resolution as
a combination of distinct problems of image registration and
sparse coding. Main contribution of this work is formulating these two
problems within a framework of double sparsity dictionary learning.
Image registration and sparse coding problems are unified in a single
objective function, then registration coefficients and sparse
coding coefficients are alternatingly optimized with 
quadratic programming and l1-norm constraint least squares,
respectively, both of which
lead sparse estimation of the coefficients.
The proposed method improved our previous formulation of
multi-frame super resolution for some images.
Particularly, images from movies are significantly improved.
We mainly explained the proposed super resolution method with an
application to image resolution enhancement, however, 
we consider that our double-sparsity formulation is applicable to enhancement or
refinement of multiple observations from a number of inaccurate sensors
with appropriate base dictionaries, such as information integration from 
observations by autonomous mobile robots.
 Our future work includes application of the proposed method to other signal resolution
enhancement, and further improvement of computational efficiency by
adopting or developing optimization methods for sparse coding and shift estimation.

 \section*{Acknowledgment}
Part of this work was supported by JSPS KAKENHI No. 25120009 and 26120504.

\end{document}